\documentclass{article}
\usepackage[utf8]{inputenc} 
\usepackage[T1]{fontenc}    
\usepackage{hyperref}       
\hypersetup{colorlinks=true,linkcolor=blue,citecolor=blue}
\usepackage{url,fullpage} 
\usepackage{booktabs}       
\usepackage{amsfonts}       
\usepackage{nicefrac}       
\usepackage{microtype}      
\usepackage{pifont,xspace,epsfig,wrapfig,paralist}
\usepackage{amsmath,amsthm,amssymb,times}
\theoremstyle{plain}
\usepackage{latexsym,paralist,multirow,tablefootnote}
\usepackage{graphicx}
\usepackage{amsfonts}
\usepackage{makecell}
\usepackage{graphicx}
\usepackage{adjustbox}
\usepackage{subcaption}
\newcounter{magicrownumbers}

\newenvironment{CompactEnumerate}{
\begin{list}{\arabic{enumi}.}{%
\usecounter{enumi}
\setlength{\leftmargin}{0pt}
\setlength{\itemindent}{0pt}
\setlength{\topsep}{0ex}
\setlength{\itemsep}{0ex}
}}
{\end{list}}

\newtheorem{theorem}{Theorem}[section]
\newtheorem*{non-theorem}{Theorem}

\newtheorem{proposition}[theorem]{Proposition}

\newtheorem{definition}{Definition}

\def \SK {\textsc{Sk}\xspace}

\def \FC {\textsc{Fc}\xspace}
\def \Conv {\textsc{Conv}\xspace}

\def \Loss {\mathrm {Loss}}

\def \inp {\mathrm {in}}
\def \out {\mathrm {out}}
\def \mat {\mathrm {mat}}

\def \z {\mathbf z}

\def \E {\mathbb{E}}

\def \u {\mathbf u}

\def \a {\mathbf a}
\def \b {\mathbf b}
\def \h {\mathbf h}

\def \g {\mathbf g}
\def \m {\mathbf m}

\def \v {\mathbf v}
\def \z {\mathbf z}
\def \w {\mathbf w}

\def \R {\mathbb{R}}

\newcommand{\NN}{\mbox{\sc NN}}
\newcommand{\LOWRANK}{\mbox{\sc LowRank}}
\newcommand{\MAXPOOL}{\mbox{\sc MaxPool}}

\newcommand{\topone}{\mbox{\sc ErrTop-1}}

\begin{document}
\title{Deep Neural Network Approximation using Tensor Sketching}
\author{Shiva Prasad Kasiviswanathan\thanks{Amazon ML. Work done while the author was at Samsung Research America, Mountain View, CA, USA. \href{mailto:kasivisw@gmail.com}{kasivisw@gmail.com}.}\;\,\thanks{Equal Contributions.} \and Nina Narodytska\thanks{VMware Research, Palo Alto, CA, USA. \href{mailto:n.narodytska@gmail.com}{n.narodytska@gmail.com}.}\;\,\footnotemark[2]\and Hongxia Jin\thanks{Samsung Research America, Mountain View, CA, USA. \href{mailto:hongxia.jin@samsung.com}{hongxia.jin@samsung.com }.}}
\date{}
\maketitle

\begin{abstract}
Deep neural networks are powerful learning models that achieve state-of-the-art performance on many computer vision, speech, and language processing tasks. In this paper, we study a fundamental question that arises when designing deep network architectures: Given a target network architecture can we design a ``smaller'' network architecture that ``approximates'' the operation of the target network? The question is, in part, motivated by the challenge of parameter reduction (compression) in modern deep neural networks, as the ever increasing storage and memory requirements of these networks pose a problem in resource constrained environments. 

In this work, we focus on deep convolutional neural network architectures, and propose a novel randomized tensor sketching technique that we utilize to develop a unified framework for approximating the operation of both the convolutional and fully connected layers. By applying the sketching technique along different tensor dimensions, we design changes to the convolutional and fully connected layers that substantially reduce the number of effective parameters in a network. We show that the resulting smaller network can be trained directly, and has a classification accuracy that is comparable to the original network. 
\end{abstract}

\section{Introduction} \label{sec:intro}
Deep neural networks have become ubiquitous in machine learning with applications, ranging from computer vision, to speech recognition, and natural language processing.  The recent successes of convolutional neural networks (CNNs) for computer vision applications have, in part, been enabled by recent advances in scaling up these networks, leading to networks with millions of parameters. As these networks keep growing in their number of parameters, reducing their storage and computational costs has become critical for meeting the requirements of practical applications. Because while it is possible to train and deploy these deep convolutional neural networks on modern clusters, their storage, memory bandwidth, and computational requirements make them prohibitive for embedded mobile applications. On the other hand, computer vision applications are growing in importance for mobile platforms. This dilemma gives rise to the following natural question: {\em Given a target network architecture, is it possible to design a new smaller network architecture (i.e., with fewer parameters), which approximates the original (target) network architecture in its operations on all inputs?} In this paper, we present an approach for answering this {\em network approximation} question using the idea of {\em tensor sketching.} 

Network approximation is a powerful construct because it allows one to replace the original network with the smaller one for both training and subsequent deployment~\cite{denil2013predicting,HashedNets,cheng2015exploration,yang2015deep,sindhwani2015structured,FreshNets,tai2015convolutional,garipov2016ultimate}.\!\footnote{For clarity, we distinguish between the terms network and model in this paper: network refers to network architecture that describes the transformation applied on the input, whereas model refers to a trained network with fixed parameters obtained by training a network with some training set.} 
That is, it completely eliminates the need for ever realizing the original network, even during the initial training phase, which is a highly desirable property when working in a  storage and computation constrained environments. While approximating any network (circuit) using a smaller network (circuit) is computationally a hard problem~\cite{umans1998minimum}, in this paper, we study the problem of network approximation on convolutional neural networks. To approximate a convolutional neural network $\NN$, we focus on its parametrized layers (the convolutional and fully connected layers). Consider any such layer $L$ in the network $\NN$. Let $\phi \,: \, \Gamma \times \Theta \rightarrow \Omega$ denote the function (transformation) applied by this layer, where $\Theta$ represents the parameter space of the function (generally, a tensor space of some order), $\Gamma$ and $\Omega$ represent the input and output space respectively. Our general idea is to replace $\phi$ by a randomized function $\hat{\phi} \,:\, \Gamma \times \widehat{\Theta} \rightarrow \Omega$, such that  $\forall \theta \in \Theta, \; \exists \hat{\theta} \in \widehat{\Theta}, \; \mbox{ such that for every input} \gamma \in \Gamma, \; \E[\hat{\phi}(\gamma;\hat{\theta})] = \phi(\gamma;\theta)$,  
where the expectation is over randomness of the function $\hat{\phi}$. In other words, $\hat{\phi}(\gamma;\hat{\theta})$ is an unbiased estimator of $\phi(\gamma;\theta)$. Additionally, we establish theoretical bounds on the variance of this estimator. Ideally, we want the representation length of $\hat{\theta}$ to be much smaller than that of $\theta$. For the construction of $\hat{\phi}$, we introduce a novel randomized tensor sketching idea. The rough idea here is to create multiple sketches of the tensor space $\Theta$ by performing random linear projections along different dimensions of $\Theta$, and then perform a simple combination of these sketches. This new operation $\hat{\phi}$ defines a new layer that approximates the functionality $\phi$ of the layer $L$. Since $\hat{\phi}$ and $\phi$ have the same input and output dimensionality, we can replace the layer $L$ in the network $\NN$ with this new (sketch counterpart) layer. Doing so for all the convolutional and fully connected layers in $\NN$, while maintaining the rest of the architecture, leads to a smaller network $\widehat{\NN}$, which approximates the network $\NN$. To the best of our knowledge, ours is the first work that uses the idea of sketching of the parameter space for the task of network approximation.

The next issue is: Can we efficiently train the smaller network $\widehat{\NN}$? We show that, with some changes to the standard training procedure, the parameters (which now represent sketches) of the constructed smaller network can be learnt space efficiently on any training set. Also compared to the original network, there is also a slight improvement in the running time needed for various operations in this smaller network. This allows us to train $\widehat{\NN}$ directly on $D$ to get a reduced model $\widehat{\NN}_D$.\!\footnote{There memory footprint of the reduced model $\widehat{\NN}_D$ can be further reduced using various careful operations such as pruning, binarization, quantization, low-rank decomposition, etc.,~\cite{gong2014compressing,han2015learning,han2015deep,soulie2015compression,wu2015quantized,guo2016dynamic,kim2015compression,wang2016cnnpack,hubara2016binarized,hubara2016quantized,li2016ternary,zhu2016trained}, which is beyond the scope of this work.} Our extensive experimental evaluations, on different datasets and architectures, corroborate the excellent performance of our approach by showing that it increases the limits of achievable parameter number reduction while almost preserving the original model accuracy, compared to several existing approximation techniques.  In fact, our technique succeeds in generating smaller networks that provide good accuracy even on large datasets such as Places2, which other state-of-the-art network approximation techniques seem not to succeed on.


\subsection{Preliminaries} \label{sec:prelim}
We denote $[n]=\{1,\ldots,n\}$. Vectors are in column-wise fashion, denoted by boldface letters.  For a vector $\v$, $\v^\top$ denotes its transpose and $\| \v \|$ its Euclidean norm. For a matrix $M$, $\| M\|_F$ denotes its Froebnius norm. We use random matrices to create sketches of the matrices/tensors involved in the fully connected/convolutional layers. In this paper, for simplicity, we use random scaled sign (Rademacher) matrices. We note that other families of distributions such as subsampled randomized Hadamard transforms can probably lead to additional computational efficiency gains when used for sketching.    

\begin{definition} \label{defn:probing}
Let $Z \in \R^{k \times d}$ be a random sign matrix with independent entries that are $+1$ or $-1$ with probability $1/2$. We define a random scaled sign matrix $U = Z/\sqrt{k}$.
\end{definition}
Here, $k$ is a parameter that is adjustable in our algorithm. We generally assume $k \ll d$. Note that $\E[U^\top U ] = \mathbb{I}_{d}$ where $\mathbb{I}_d$ is the $d \times d$ identity matrix.  Also by linearity of expectation, for any matrix $M$ with $d$ columns, we have $\E[M U^\top U ] = M \E[U^\top U] = M$.

\noindent\textbf{Tensor Preliminaries.} We denote matrices by uppercase letters  and higher dimensional tensors by Euler script letters, e.g., $\mathcal{T}$. A real $p$th order tensor $\mathcal{T} \in \otimes_{i=1}^p \R^{d_i}$ is a member of the tensor product of Euclidean spaces $\R^{d_i}$ for $i \in [p]$. As is the\ case for vectors (where $p = 1$) and matrices (where $p = 2$), we may identify a $p$th order tensor with the $p$-way array of real numbers. The different dimensions of the tensor are referred to as {\em modes}. The $(i_1,\dots,i_p)$th entry of a tensor $\mathcal{T}$ is denoted by $\mathcal{T}_{i_1 i_2\dots i_p}$.

The mode-$n$ matrix product (for $n \in [p]$) of a tensor $\mathcal{T} \in \R^{d_1 \times \dots \times d_p}$ with a matrix $M \in \R^{k \times d_n}$ is denoted by $\mathcal{T} \times_{n} M$ and has dimensions $d_1 \times \dots \times d_{n-1} \times k \times d_{n+1} \times \dots d_p$. Elementwise, we have
\begin{align*}
(\mathcal{T} \times_{n} M)_{i_1\dots i_{n-1} j i_{n+1} \dots i_p} = \sum_{i_n=1}^{d_n} \mathcal{T}_{i_1 i_2 \dots i_p} M_{j i_n}.
\end{align*}
Note that the operation can also be applied simultaneously to multiple modes. In general, given $p$ matrices $M_1,\dots,M_p$ where $M_i \in \R^{k_i \times d_i}$, the resulting tensor $\mathcal{T} \times_1 {M_1} \times_{2} M_2 \dots \times_{p} M_p$ is a tensor in $\R^{k_1\times k_2 \dots \times k_p}$. For a matrix  $W \in \R^{d_1 \times d_2}$ is a matrix, it follows that: $W \times_1 M_1 = M_1 W$ and $W \times_2 M_2 = W M_2^\top$.

A \emph{fiber} of  $\mathcal{T}$ is obtained by fixing all but one of the indices of the tensor. A flattening of tensor $\mathcal{T}$ along a mode (dimension) $n$ (denoted by $\mat_n$) is a matrix whose columns correspond to mode-$n$ fibers of $\mathcal{T}$. For example, in a fourth order tensor $\mathcal{T} \in \R^{d_1 \times d_2 \times d_3 \times d_4}$, $T = \mat_4(\mathcal{T}) \in \R^{d_1d_2d_3 \times d_4}$ is a matrix defined as: $T_{(i_1 + d_1(i_2 -1 )+d_1 d_2(i_3-1))i_4} = \mathcal{T}_{i_1 i_2 i_3 i_4}$, i.e., the $(i_1,i_2,i_3,i_4)$ entry in the tensor $\mathcal{T}$ is assigned to the location $(i_1 + d_1(i_2 -1 )+d_1 d_2(i_3-1),i_4)$ in the matrix~$T$.

The weights of all (two dimensional) filters in a convolutional layer can be denoted by a $4$-dimensional tensor in $\R^{d_2 \times w \times h \times d_1}$ where $d_1$ and $d_2$ represent the number of output and input feature maps, and $h$ and $w$ represent the height and width of the filter kernels. 

\section{Tensor Sketching}  \label{sec:tensor}
Our network approximation is based on the idea of tensor sketching. Data sketching ideas have been successfully used in designing many machine-learning algorithms, especially in the setting of streaming data, see e.g.,~\cite{TCS-060}. Generally, sketching is used to construct a compact representation of the data so that certain properties in the data are (approximately) preserved. Our usage of sketching is however slightly different, instead of sketching the input data, we apply sketching on the parameters of the function. Also, we want to design sketching techniques that work uniformly for both matrices and higher order tensors. For this, we define a new tensor sketch operation defined as follows. 
\begin{definition} [Mode-$n$ Sketch] \label{defn:tensorsketch}
Given a tensor, $\mathcal{T} \in \otimes_{i=1}^p \R^{d_i}$, the mode-$n$ sketch of $\mathcal{T}$ with respect to a random scaled sign matrix $U_n \in \R^{k \times d_n}$ for $n \in [p]$, is defined as the tensor $\mathcal{S}_n = \mathcal{T} \times_n U_n$.
\end{definition}
Since, we generally pick $k \ll d_n$, the space needed for storing the sketch $\mathcal{S}_n$ is a factor $d_n/k$ smaller than that for storing $\mathcal{T}$. In the case of matrices, the sketches are created by pre- or post-multiplying the matrix with random scaled sign matrices of appropriate dimensions. For example, given a matrix $W \in \R^{d_1 \times d_2}$, we can construct mode-$1$ sketch (resp.\ mode-$2$ sketch) of $W$ as $W \times_1 U_1 = U_1 W$ (resp.\ $W \times_2 \,U_2 = W U_2^\top$). Given a sketch $S_1 = W \times_1 U_1$ (resp.\ $S_2 = W \times_2 U_2$) of a matrix $W$ and another matrix $M \in \R^{d_2 \times d_3}$, it is natural to use $U_1^\top S_1 M$ (resp.\ $S_2 U_2 M$) as an estimator for the matrix product $W M$. It is easy to see that both these estimators are unbiased. The second part of the following proposition (proof in Appendix~\ref{app:tensor}) analyzes the variance of these estimators. The result will motivate our construction of sketch-based  convolutional and fully connected layers in the next section. 
\begin{proposition} \label{prop:prop}
Let $W \in \R^{d_1 \times d_2}$. Let $U_1 \in \R^{k \times d_1}$ and $U_2 \in \R^{k \times d_2}$ be two independent random scaled sign matrices. Let $S_1 = U_1 W (= W \times_1 U_1) $ and $S_2 =W U_2^\top (= W \times_2 U_2)$. Then for any matrix $M \in \R^{d_2 \times d_3}$:
\begin{CompactEnumerate}
\item  $\E[U_1^\top S_1 M ] = W M, \mbox{ and } \; \E[S_2 U_2 M] = W M.$
\item $\E \left [ \left \| U_1^\top S_1 M - W M  \right \|_F^2 \right] \leq \frac{2 d_1\|W M\|_F^2 }{k}, \mbox{ and}$
\item[] $\;\;\; \E \left [ \left \| S_2 U_2 M - W M \right \|_F^2 \right] \leq \frac{2\|W\|_F^2 \| M \|_F^2}{k}.$
\end{CompactEnumerate}
\end{proposition}
Notice that the variance terms decrease as $1/k$. The variance bound can also be plugged into Chebyshev's inequality to get a probability bound. Also the variance bounds are quantitatively different based on whether the sketch $S_1$ or $S_2$ is used. In particular, depending on $W$ and $M$, one of the variance bounds could be substantially smaller than the other one, e.g., if the columns in $M$ are in the null space of $W$ then $WM$ is a zero matrix, so while one bound gives a tight zero variance the other one does not. 

\section{Sketch-based Network Architecture} \label{sec:architecture}
We now describe our idea of approximating a network using tensor sketching. Our approach, in almost identical fashion, can be used to reduce the number of parameters involved in both the convolutional and the fully connected layers without significantly affecting the resulting accuracy. 
\subsection{Sketching Convolutional Layers} 
A typical convolutional layer in a CNN transforms a $3$-dimensional input tensor $\mathcal{I}_{\inp} \in \R^{h_1 \times w_1 \times d_2}$ into a output tensor $\mathcal{I}_{\out} \in \R^{h_2 \times w_2 \times d_1}$ by convolving $\mathcal{I}_{\inp}$ with the kernel tensor $\mathcal{K} \in \R^{d_2 \times h \times w \times d_1}$, where $h_2$ and $w_2$ depends on $h,w,h_1,w_1$ and possibly other parameters such as stride, spatial extent, zero padding~\cite{Goodfellow-et-al-2016-Book}. We use $\ast$ to denote the convolution operation, $\mathcal{I}_\out = \mathcal{I}_\inp \ast \mathcal{K}$.
The exact definition of the convolution operator ($\ast$) that depends on these above mentioned additional parameters is not very important for us, and we only rely on the fact that the convolution operation can be realized using a matrix multiplication as we explain below.\!\footnote{In a commonly used setting, with stride of $1$ and zero-padding of $0$, $h_2 =h_1-h+1$ and $w_2 = w_1 - w + 1$, and $\mathcal{I}_{\out} \in \R^{(h_1-h+1) \times (w_1 -w +1) \times d_1}$ is defined as:
$\mathcal{I}_{\out_{xys}} = \sum_{i=1}^{h} \sum_{j=1}^w \sum_{c=1}^{d_2} \mathcal{K}_{cijs} \; \mathcal{I}_{\inp_{(x+i-1)(y+j-1)c}}$.}
Also a convolutional layer could be optionally followed by application of some non-linear activation function (such as ReLU or tanh), which are generally parameter free, and do not affect our construction.  

\begin{figure*}  
\begin{center}
\begin{tabular}{cc}
\begin{subfigure}{\textwidth}\centering\includegraphics[width=6in]{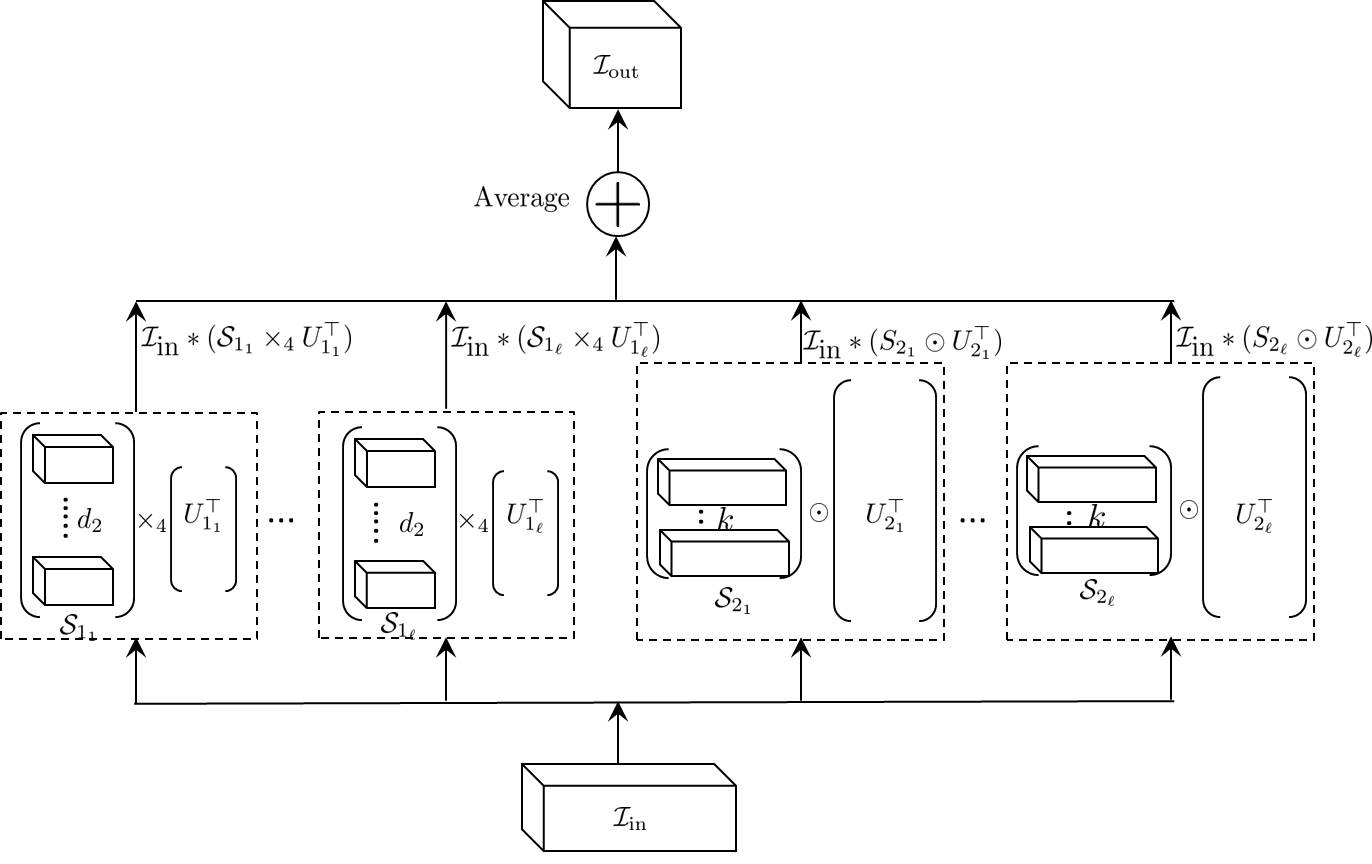}\end{subfigure}
\end{tabular} 
\vspace*{-1ex}
\caption{A \SK-\Conv layer with parameters $(\mathcal{S}_{1_1},U_{1_1}),\dots,(\mathcal{S}_{1_\ell},U_{1_\ell})$, $(\mathcal{S}_{2_1},U_{2_1}),\dots,(\mathcal{S}_{2_\ell},U_{2_\ell})$.} 
\label{fig:skconv}
\end{center}
\end{figure*}

We use the tensor sketch operation (Definition~\ref{defn:tensorsketch}) to reduce either the dimensionality of the input feature map ($d_2$) or the output feature map ($d_1$) in the kernel tensor $\mathcal{K}$. In practice, the dimensions of the individual filters ($h$ and $w$) are small integers, which we therefore do not further reduce. The motivation for sketching along different dimensions comes from our mathematical analysis of the variance bounds (Theorem~\ref{thm:convprop}), where as in Proposition~\ref{prop:prop} based on the relationship between $\mathcal{I}_\inp$ and $\mathcal{K}$ the variance could be substantially smaller in one case or the other. Another trick that works as a simple boosting technique is to utilize multiple sketches each associated with an independent random matrix. Formally, we define a \SK-\Conv layer as follows (see also Figure~\ref{fig:skconv}).

\begin{definition} \label{defn:conv}
A \SK-\Conv layer is parametrized by a sequence of tensor-matrix pairs $(\mathcal{S}_{1_1},U_{1_1}),\dots,(\mathcal{S}_{1_\ell},U_{1_\ell})$, $(\mathcal{S}_{2_1},U_{2_1}),\dots,(\mathcal{S}_{2_\ell},U_{2_\ell})$ where for $i \in [\ell]$ $\mathcal{S}_{1_i} \in \R^{d_2 \times h \times w \times k}$, $\mathcal{S}_{2_i} \in \R^{k \times h \times w \times d_1}$ and $U_{1_i} \in \R^{k \times d_1}$, $U_{2_i} \in \R^{k h w \times d_2 h w}$ are independent random scaled sign matrices,\footnote{We define $U_{2_i} \in \R^{k h w \times d_2 h w}$ (instead of $U_{2_i} \in \R^{k \times d_2 }$) for simplifying the construction.} which on input $\mathcal{I}_{\inp} \in \R^{h_1 \times w_1 \times d_2}$ constructs $\mathcal{I}_\out$ as follows:
\begin{align} \label{eqn:convsketch}
\hspace*{-1ex}
\mathcal{I}_\out = \frac{1}{2\ell} \sum_{i=1}^\ell \mathcal{I}_{\inp} \ast (\mathcal{S}_{1_i} \times_4 U_{1_i}^\top) + \frac{1}{2\ell} \sum_{i=1}^\ell \mathcal{I}_{\inp} \ast (\mathcal{S}_{2_i} \odot U_{2_i}^\top),
\end{align}
where $\mathcal{S}_{2_i} \odot U_{2_i}^\top \in \R^{d_2 \times h \times w \times d_1}$ is defined as\footnote{Let $\mathcal{O}_i = \mathcal{S}_{2_i} \odot U_{2_i}^\top$. The $\odot$ operation can be equivalently defined: $\mat_4(\mathcal{O}_i) = U_{2_i}^\top \mat_4(\mathcal{S}_{2_i})$.} 
\begin{align*}(\mathcal{S}_{2_i} \odot U_{2_i}^\top)_{xyzs} = \sum_{c=1}^{k}\sum_{i=1}^{h} \sum_{j=1}^{w} \mathcal{S}_{{2_i}_{c i j s}} U_{{2_i}_{(cij)(xyz)}}.\end{align*}
Here $(\mathcal{S}_{2_i} \odot U_{2_i}^\top)_{xyzs}$ is the $(x,y,z,s)$th entry,  $\mathcal{S}_{{2_i}_{c i j s}}$ is the $(c,i,j,s)$th entry, and $U_{{2_i}_{(cij)(xyz)}}$ is the $(cij,xyz)$th entry in $(\mathcal{S}_{2_i} \odot U_{2_i}^\top)$, $\mathcal{S}_{{2_i}}$, and  $U_{{2_i}}$, respectively.
\end{definition}
By running multiple sketches in parallel on the same input and taking the average, also results in a more stable performance across different choices of the random matrices (see the experimental discussion in Appendix~\ref{app:exp}). The number of free parameters overall in all the $\mathcal{S}_{1_i}$ and $\mathcal{S}_{2_i}$  tensors put together equals $\ell hw k ( d_1 + d_2)$.\!\footnote{The random matrices, once picked are not changed during the training or deployment.}   Therefore, with a  \SK-\Conv layer, we get a reduction in the number of parameters compared to a traditional convolutional layer (with $h w d_1 d_2$ parameters) if $k \ell  \leq d_1 d_2/(d_1+d_2)$. With this reduction, the time for computing $I_\out$, ignoring dependence on $h$ and $w$, reduces from $O(h_2w_2 d_1 d_2)$ (in a traditional \Conv layer) to $O(h_2w_2 \ell k (d_1+d_2))$ (in a \SK-\Conv layer).

The convolution operation can be reduced into a matrix multiplication, an idea that is exploited by many deep learning frameworks~\cite{chetlur2014cudnn}. The idea is to reformulate the kernel tensor $\mathcal{K}$ by flattening it along the dimension representing the output feature map, which in our setting is represented along the fourth dimension of $\mathcal{K}$. The input tensor $\mathcal{I}_{\inp}$ is used to form a matrix $I_{\inp} \in \R^{h_2w_2 \times d_2 h w}$. This construction is quite standard and we refer the reader to~\cite{chetlur2014cudnn} for more details. Then it follows that $I_\out$ defined as  $I_\inp \mat_4(\mathcal{K}) \in \R^{h_2 w_2 \times d_1}$ is a reshaping of the output tensor $\mathcal{I}_\out$ (i.e., $I_\out = \mat_3(\mathcal{I}_\out)$).

Using this equivalence and simple algebraic observations ($\mat_4(\mathcal{S}_{1_i} \times_4 U_{1_i}^\top) = \mat_4(\mathcal{S}_{1_i}) U_{1_i}$ and $\mat_4(\mathcal{S}_{2_i} \odot U_{2_i}^\top) = U_{2_i}^\top \mat_4(\mathcal{S}_{2_i})$), we can re-express the operation in~\eqref{eqn:convsketch} as:
\begin{align} \label{eqn:newconvsketch} 
I_\out = \frac{1}{2\ell} \sum_{i=1}^\ell I_{\inp} \mat_4(\mathcal{S}_{1_i}) U_{1_i} + \frac{1}{2\ell} \sum_{i=1}^\ell I_{\inp} U_{2_i}^\top \mat_4(\mathcal{S}_{2_i}).
\end{align}
Or in other words,
\begin{align*}
I_\out = \frac{1}{2\ell} \sum_{i=1}^\ell I_{\inp} (\mat_4(\mathcal{S}_{1_i}) \times_2 U_{1_i}^\top) + \frac{1}{2\ell} \sum_{i=1}^\ell I_{\inp} (\mat_4(\mathcal{S}_{2_i}) \times_1 U_{2_i}^\top) .  
\end{align*} 

\noindent\textbf{Theoretical Guarantees of a \SK-\Conv Layer.} Given a traditional convolutional layer with kernel tensor $\mathcal{K}$ and independent random scaled sign matrices $U_{1_1},\dots,U_{1_\ell},$  $U_{2_1},\dots,U_{2_\ell}$, we can form a corresponding \SK-\Conv layer by constructing tensors $\mathcal{S}_{1_1},\dots,\mathcal{S}_{1_\ell},\mathcal{S}_{2_1},\dots,\mathcal{S}_{2_\ell}$ such that $\mat_4(\mathcal{S}_{1_i}) = \mat_4(\mathcal{K}) U_{1_i}^\top$ and $\mat_4(\mathcal{S}_{2_i}) = U_{2_i} \mat_4(\mathcal{K})$ for $i \in [\ell]$. The following theorem (proof in Appendix~\ref{app:convprop}), based on Proposition~\ref{prop:prop}, analyzes the expectation and the variance of using these sketches as an estimator for $\mathcal{I}_\inp \ast \mathcal{K}$ ($\equiv I_\inp \mat_4(\mathcal{K})$). Since the random matrices are independent of each other, we drop the subscript  and perform the analysis for a single instantiation of these sketches. .  
\begin{theorem} \label{thm:convprop}
Let $\mathcal{K} \in \R^{d_2 \times h \times w \times d_1}$ be a kernel tensor and $K = \mat_4(\mathcal{K})$. Let $U_1 \in \R^{k \times d_1}$ and $U_2 \in \R^{k h w \times d_2 h w}$ be two independent random scaled sign matrices. Let $\mathcal{S}_1$ and $\mathcal{S}_2$ be tensors such that $\mat_4(\mathcal{S}_1) = K \times_2 U_1$ and $\mat_4(\mathcal{S}_2) = K \times_1 U_2$. Then for any input matrix $I_\inp \in \R^{h_2 w_2 \times d_2 h w}$ (formed from an input tensor $\mathcal{I}_\inp \in \R^{h_1 \times w_1 \times d_2}$):
\begin{CompactEnumerate}
\item Unbiased Estimation: $\E[I_\inp\, \mat_4(\mathcal{S}_1) U_1] = I_\inp K$, and  $\E[I_\inp U_2^\top \, \mat_4(\mathcal{S}_2)] = I_\inp K.$
\item Variance Bound:
\vspace*{-1ex} 
\begin{align*}
& \E  \left [ \left \| I_\inp \, \mat_4(\mathcal{S}_1) U_1 - I_\inp K \right \|_F^2 \right ] \leq \frac{2 d_1\|I_\inp K\|_F^2 }{k}, \mbox{ and} \\
&\E \left [ \left \| I_\inp U_2^\top \, \mat_4(\mathcal{S}_2) - I_\inp K \right \|_F^2 \right ] \leq \frac{2 \|I_\inp\|_F^2 \| K \|_F^2}{k h w}.
\end{align*}
\end{CompactEnumerate}
\end{theorem}

\subsubsection{Training a \SK-\Conv Layer}  
In this section, we discuss a procedure for training a \SK-\Conv layer. Let $\Loss()$ denote some loss function for the network.  For computational and space efficiency, our goal will be to perform the training without ever needing to construct the complete kernel tensor ($\mathcal{K}$). We focus on deriving the gradient of the loss with respect to the parameters in a \SK-\Conv layer, which can then be used for back-propagating the gradient information.

We can again exploit the equivalence between the convolution operation and matrix multiplication. Consider the operation performed in the \SK-\Conv layer as defined in~\eqref{eqn:newconvsketch}.
Let $G = \frac{\partial \Loss}{\partial I_\out} \in \R^{h_2 w_2 \times d_1}$. For $i \in [\ell]$,\footnote{The gradients computed with respect to $\mat_4(\mathcal{S}_{1_i})$ and $\mat_4(\mathcal{S}_{2_i})$ can also be converted into a tensor by reversing the $\mat_4()$ operator.}
\begin{eqnarray*}
&\frac{\partial \Loss}{\partial\;  \mat_4(\mathcal{S}_{1_i})} = \frac{I_\inp^\top G U_{1_i}^\top}{2\ell},& \\
&\frac{\partial \Loss}{\partial\; \mat_4(\mathcal{S}_{2_i})} = \frac{U_{2_i} I_\inp^\top G}{2\ell}, \mbox{ and}& \\
&\frac{\partial \Loss}{\partial  I_\inp} = \sum_{i=1}^\ell \frac{G U_{1_i}^\top \; \mat_4(\mathcal{S}_{1_i})^\top}{2 \ell} + \sum_{i=1}^\ell \frac{G \; \mat_4(\mathcal{S}_{2_i})^\top U_{2_i}}{2 \ell}.&
\end{eqnarray*}
Notice that all the required operations can be carried out without ever explicitly forming the complete $d_2 \times h \times w \times d_1$ sized kernel tensor.

\subsection{Sketching Fully Connected Layers} \label{sec:sketchfc}
Neurons in a fully connected (\FC) layer have full connections to all activations in the previous layer. These layers apply a linear transformation of the input. Let $W \in \R^{d_1 \times d_2}$ represent a {\em weight} matrix and $\b \in \R^{d_1}$ represent a {\em bias} vector. The operation of the \FC layer on input $\h_\inp$ can be described as:
\begin{align} \label{eqn:stdforward}
\a = W \h_{\inp} + \b. 
\end{align}
Typically, the \FC layer is followed by application of some non-linear activation function. As in the case of convolutional layers, our construction is independent of the applied activation function and we omit further discussion of these functions.

Our idea is to use the tensor sketch operation (Definition~\ref{defn:tensorsketch}) to sketch either the columns or rows of the weight matrix. 

\begin{definition} \label{defn:fc}
A \SK-\FC layer is parametrized by a bias vector $\b \in \R^{d_1}$ and a sequence of matrix pairs $(S_{1_1},U_{1_1}),\dots$, $(S_{1_\ell},U_{1_\ell}),(S_{2_1},U_{2_1}),\dots,(S_{2_\ell},U_{2_\ell})$ where for $i \in [\ell]$, $S_{1_i} \in \R^{k \times d_2}$, $S_{2_i} \in \R^{d_1 \times k}$ and $U_{1_i} \in \R^{k \times d_1}$, $U_{2_i} \in \R^{k \times d_2}$ are independent random scaled sign matrices, which on input $\h_{\inp} \in \R^{d_2}$ performs the following operation:
\begin{align} \label{eqn:fcsketch}
& \a = \frac{1}{2\ell} \sum_{i=1}^\ell U_{1_i}^\top S_{1_i} \h_{\inp} + \frac{1}{2\ell} \sum_{i=1}^\ell S_{2_i} U_{2_i} \h_{\inp} + \b.
\end{align}
\end{definition}
Note that $\a$ in the above definition could be equivalently represented as:
 \begin{align*}
 \a = \frac{1}{2\ell} \sum_{i=1}^\ell (S_{1_i} \times_1 U_{1_i}^\top)\h_{\inp} + \frac{1}{2\ell} \sum_{i=1}^\ell (S_{2_i} \times_2 U_{2_i}^\top) \h_{\inp} + \b.
\end{align*}
The number of free parameters overall in all the $S_{1_i}$ and $S_{2_i}$ matrices put together is $\ell k (d_1 + d_2)$. Therefore, compared to a traditional weight matrix $W \in \R^{d_1 \times d_2}$, we get a reduction in the number of parameters if $k \ell \leq d_1 d_2/(d_1+d_2)$. Another advantage is that the time needed for computing the pre-activation value ($\a$ in~\eqref{eqn:fcsketch}) in a \SK-\FC layer is $O(\ell k (d_1 + d_2))$ which  is smaller than the $O(d_1 d_2)$ time needed in the traditional \FC setting if the values of $k$ and $\ell$ satisfy the above condition.


\noindent\textbf{Theoretical Guarantees of \SK-\FC Layer.} Given a traditional \FC layer with weight matrix $W$ (as in~\eqref{eqn:stdforward}), and independent  random scaled sign matrices $U_{1_1},\dots,U_{1_\ell},U_{2_1},\dots,U_{2_\ell}$, we can form a corresponding \SK-\FC layer by setting $S_{1_i} = U_{1_i} W$ and $S_{2_i} = W U_{2_i}^\top$. We now analyze certain properties of this construction.  The following theorem, based on Proposition~\ref{prop:prop}, analyzes the expectation and the variance of using these sketches as an estimator for $W \h_\inp + \b$ for a vector $\h_\inp \in \R^{d_2}$. Since the random matrices are independent of each other, we drop the subscript and perform the analysis for a single instantiation of these sketches. 

\begin{theorem} 
Let $W \in \R^{d_1 \times d_2}$. Let $U_1 \in \R^{k \times d_1}$ and $U_2 \in \R^{k \times d_2}$ be two independent random scaled sign matrices. Let $S_1 = U_1 W (= W \times_1 U_1) $ and $S_2 =W U_2^\top (= W \times_2 U_2)$. Then for any $\h_{\inp} \in \R^{d_2}$ and $\b \in \R^{d_1}$:
\begin{CompactEnumerate}
\item \label{parta} Unbiased Estimation:  $\E[U_1^\top S_1 \h_\inp + \b] = W \h_\inp + \b$, and $\E[S_2 U_2 \h_\inp + \b] = W \h_\inp + \b.$
\item \label{partb} Variance Bound: 
\begin{align*}
& \E \left [ \left \| U_1^\top S_1 \h_\inp + \b - (W \h_\inp + \b) \right \|^2 \right ] \leq \frac{2 d_1\|W \h_\inp\|^2 }{k}, \\  
&\E \left [ \left \| S_2 U_2 \h_\inp + \b - (W \h_\inp + \b) \right \|^2 \right] \leq \frac{2\|W \|_F^2 \| \h_\inp \|^2}{k}.
\end{align*}
\end{CompactEnumerate}
\end{theorem}


\begin{figure*}  
\begin{center}
\begin{tabular}{cc}
\begin{subfigure}{\textwidth}\centering\includegraphics[width=6in]{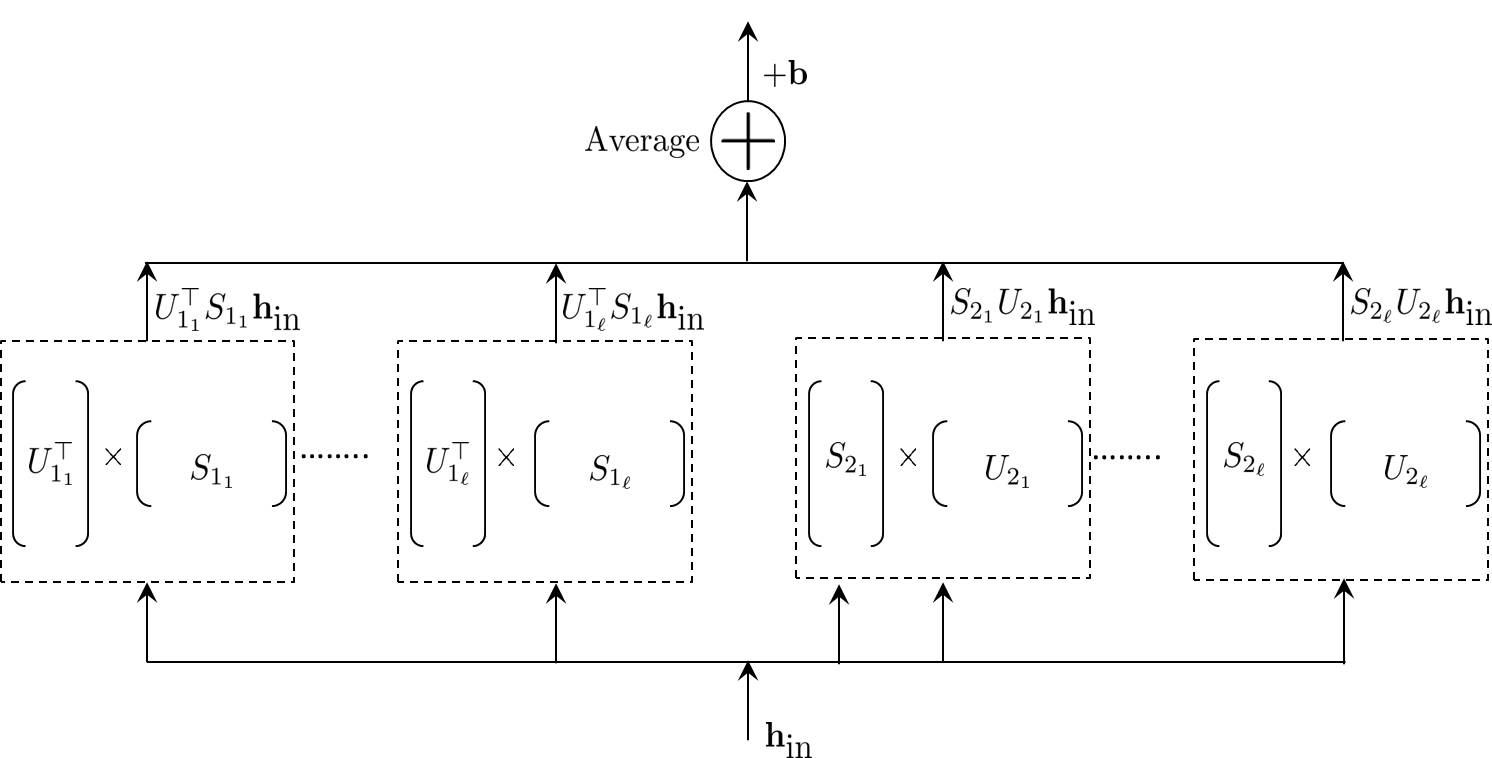}\end{subfigure}
\end{tabular} 
\caption{A \SK-\FC layer with parameters $\b$, $(S_{1_1},U_{1_1}),\dots$, $(S_{1_\ell},U_{1_\ell}),(S_{2_1},U_{2_1}),\dots,(S_{2_\ell},U_{2_\ell})$.} 
\label{fig:skfc}
\end{center}
\end{figure*}

\subsubsection{Training a \SK-\FC Layer} \label{sec:training}
In this section, we discuss a procedure for training a network containing \SK-\FC layers. Let $\Loss()$ denote some loss function for the network.  Let $\a = S_2 U_2 \h_\inp + \b$. Let $\g = \frac{\partial  \Loss}{\partial \a}$.  In this case, using chain-rule of calculus
\begin{align} \label{eqn:1}
\frac{\partial \Loss}{\partial  S_2} = \g \h_{\inp}^\top U_2^\top = (\g \h_{\inp}^\top) \times_2 U_2.
\end{align}
Similarly, the gradient with respect to $\h_\inp$ can be calculated as:
\begin{align} \label{eqn:0}
\frac{\partial \Loss}{\partial  \h_\inp} = (S_2 U_2)^\top \g = (S_2 \times_2 U_2^\top)^\top \g. 
\end{align}

Now let $\a = U_1^\top S_1 \h_\inp + \b = (S_1^\top U_1)^\top \h_\inp + \b$. Again let $\g = \frac{\partial  \Loss}{\partial \a}$. Applying chain-rule gives 
$$\frac{\partial \Loss}{\partial  S_1} = \sum_{i=1}^{d_1} \frac{\partial \Loss}{\partial  a_i} \frac{\partial a_i}{\partial  S_1},$$ 
where $a_i$ denotes the $i$th entry of $\a$. We can compute $\frac{\partial a_i}{\partial  S_1}$ as: 
\begin{align*}
\frac{\partial a_i}{\partial  S_1} = \frac{\partial \; \u_{1_i}^\top S_1 \h_\inp }{\partial  S_1} = \u_{1_i} \h_\inp^\top,
\end{align*}
where $\u_{1_i}$ is the $i$th column in $U_1$. Therefore, we get
\begin{align} \label{eqn:2}
\frac{\partial \Loss}{\partial  S_1} = \sum_{i=1}^{d_1} g_i \u_{1_i} \h_\inp^\top = U_1 \g \h_\inp^\top = (\g \h_{\inp}^\top) \times_1 U_1,
\end{align}
where $g_i$ denotes the $i$th entry of $\g$. Finally, the gradient with respect to $\h_\inp$ in this case equals:
\begin{align} \label{eqn:3}
\frac{\partial \Loss}{\partial  \h_\inp} = (S_1^\top U_1) \g = (S_1 \times_1 U_1^\top)^\top \g.
\end{align}

Putting together~\eqref{eqn:1},~\eqref{eqn:0},~\eqref{eqn:2}, and~\eqref{eqn:3} gives the necessary gradients for the \SK-\FC layer (where $\a$ is defined using~\eqref{eqn:fcsketch}). Let $\g = \frac{\partial  \Loss}{\partial \a}$. For $i \in [\ell]$,
\begin{eqnarray*}
&\frac{\partial \Loss}{\partial  S_{1_i}} = \frac{U_{1_i} \g \h_\inp^\top}{2\ell},& \\
&\frac{\partial \Loss}{\partial  S_{2_i}} = \frac{\g \h_{\inp}^\top U_{2_i}^\top}{2\ell}, \mbox{ and}& \\
&\frac{\partial \Loss}{\partial  \h_\inp} = \sum_{i=1}^\ell \frac{S_{1_i}^\top U_{1_i} \g}{2\ell} + \sum_{i=1}^\ell \frac{U_{2_i}^\top S_{2_i} \g}{2 \ell}.&
\end{eqnarray*}
Note that all the above computations can be performed without ever explicitly forming the complete $d_1 \times d_2$ weight matrix.

\subsection{Final Construction of $\widehat{\NN}$} 
Given a convolutional neural network $\NN$, construct $\widehat{\NN}$, an approximation of $\NN$, by replacing the convolutional layers (resp.\ fully connected layers)  with \SK-\Conv layers (resp.\ \SK-\FC layers). A nice feature about this construction is that, based on need, we can also choose to replace only some of the layers of the $\NN$ with their sketch counterpart layers. 


\section{Comparison to Previous Work} \label{sec:related}

Deep neural networks are typically over-parametrized, and there is significant redundancy in deep learning networks~\cite{denil2013predicting}. There have been several previous attempts to reduce the complexity of deep neural networks under a variety of contexts. 


\noindent\textbf{Approximating only the Fully Connected Layers.} A set of techniques have focused on approximating only the fully connected layers in some reduced form. Yang~\emph{et al.}\ \cite{yang2015deep} use the {\em Fastfood} transformation technique of~\cite{le2013fastfood} to approximate the fully connected layers. The HashedNets architecture, proposed by Chen~\emph{et al.}\ \cite{HashedNets}, uses a hash function to enforce parameter sharing between random groups of parameters in a fully connected layer to reduce the number of effective parameters. Cheng~\emph{et al.}\ \cite{cheng2015exploration} achieve parameter reduction by imposing a circulant matrix structure on fully connected layers. Sindhwani~\emph{et al.}\  \cite{sindhwani2015structured} generalize this construction by proposing a broad family of structured parameter matrix structure and showing its effectiveness on the fully connected layers. Choromanska~\emph{et al.}\ \cite{choromanska2016binary} provide some theoretical justifications for using structured hashed projections in these layers. While some of these techniques are highly effective on the fully connected layers, they fall short of  achieving a significant reduction in the number parameters for modern CNNs which are dominated by convolutional layers~\cite{szegedy2015going,he2015deep}.\!\footnote{Some recent studies~\cite{long2015fully,nin} have suggested that removing fully connected layers and replacing them with convolutions and pooling could be beneficial for certain computer vision applications.} Therefore, any {\em effective} technique for parameter reduction on CNNs should {\em also} act on convolutional layers. 

\noindent\textbf{Approximating both the Convolutional and Fully Connected Layers.} Most relevant to our paper is a line of work on approximating both the fully connected and convolutional layers. Denil~\emph{et al.}\ \cite{denil2013predicting}, suggested an approach based on learning a low-rank factorization of the matrices (tensors are viewed as a matrix) involved within each layer of a CNN. Instead of learning both the factors of a factorization during training, the authors suggest techniques for carefully constructing one of the factors (called the dictionary), while only learning the other one. Our sketching-based approach is related to low-rank factorization, however using sketching we eliminate the overhead of carefully constructing the dictionary. Tai~\emph{et al.}\ \cite{tai2015convolutional} achieve parameter reduction using a tensor decomposition technique that is based on replacing the convolutional kernel with two consecutive kernels with a lower rank. The issue with this approach is that with the increased depth of the resulting network, training becomes more challenging, and the authors rely on {\em batch normalization} (proposed by~\cite{ioffe2015batch}) to overcome this issue. In our proposed approach, the depth of the reduced network remains equal to that of the original network, and the reduced network can be trained with or without batch normalization. Very recently, Garipov~\emph{et al.}\ \cite{garipov2016ultimate}, building upon a work by~\cite{novikov2015tensorizing}, used a tensor factorization technique, called tensor train decomposition, to uniformly approximate both the fully connected and convolutional layers. However, constructing an exact tensor factorization (even computing the tensor rank) is in general a challenging NP-hard problem, whereas our approach relies only on simple linear transformations. Chen~\emph{et al.}\ \cite{FreshNets} combine the hashing idea from~\cite{HashedNets} along with the discrete cosine transform (DCT) to compress filters in a convolutional layer. Their architecture, called FreshNets, first converts filter weights into frequency domain using discrete cosine transform and then uses the hashing idea to randomly group the resulting frequency parameters into buckets. Our sketches are created by using random projections which is related to the hashing trick used in these results, however, our techniques are naturally attractive for convolutional neural networks as they are known to be preserve spatial locality~\cite{JLT}, a property that is not preserved by simple hashing. Also, in contrast to FreshNets, our architectures require just simple linear transformations for both fully connected and convolutional layers, and do not require special routines for DCT, Inverse DCT, etc. Additionally, we provide theoretical bounds on the quality of approximation that is missing in these previous studies. 

\noindent\textbf{Other Related Work.} There is a long line of work on reducing model memory size based on post-processing a trained network (with sometimes further fine-tuning of the compressed model)~\cite{gong2014compressing,han2015learning,han2015deep,soulie2015compression,wu2015quantized,guo2016dynamic,kim2015compression,wang2016cnnpack,hubara2016binarized,hubara2016quantized,zhu2016trained,li2016ternary}. Techniques such as pruning, binarization, quantization, low-rank decomposition, etc., are intermingled with training of a network on a dataset to construct a reduced model. These results do not achieve a direct network approximation as the training happens on the original network. In practice, one can combine our approach with some of the above proposed model post-processing techniques to further reduce the storage requirements of the trained model (which is beyond the scope of this paper).  

Hinton~\emph{et al.}\ \cite{hinton2015distilling} and Ba~\emph{et al.}\ \cite{ba2014deep} proposed approaches to learn a ``distilled'' model, training a more compact neural network to reproduce the output of a larger network. The general idea is to train a large network on the original training labels, then learn a much smaller distilled model on a weighted combination of the original labels and the softmax output of the larger model.  Note that with our network approximation approach, we do not need to train the original large network. Also unlike distillation-based approaches where a separate distilled model has to be formed with each dataset, our approach produces a single reduced network that can be then trained on any dataset.  

Other techniques proposed for parameter reduction include inducing zeros in the parameter matrices via sparsity regularizers~\cite{collins2013memory} and storing weights in low fixed-precision formats~\cite{gupta2015deep,courbariaux2014low}. These ideas can be readily incorporated with our approach, potentially yielding further reductions in the model memory size. Daniely~\emph{et al.}\ \cite{daniely2016sketching} generate sketches of the input and show that it can lead to compact neural networks. Our approach, based on sketching the parameters of the deep network, is complementary to this idea, and the two approaches can be used in conjunction.

Several works apply related approaches to speed up the evaluation time with CNNs~\cite{jaderberg2014speeding,denton2014exploiting,lebedev2014speeding,figurnov2015perforatedcnns}. The focus of this line of work is not on parameter reduction but rather decreasing the evaluation time during testing. In each of these results, any resulting storage reduction comes as a side effect. Other techniques for speeding up convolutional neural networks include use of Winograd or FFT-based convolutions~\cite{lavin2015fast,mathieu2013fast,vasilache2014fast}. Again, unlike here, parameter reduction is not a focus of these results.

\newcommand{\TIMES}{{\mkern-2mu\times\mkern-2mu}}
\section{Experimental Evaluation} \label{sec:exp}
In this section, we experimentally demonstrate the effectiveness of our proposed network approximation approach. 
Our goal through the experiments is {\em not to test the limits of reduction} possible in deep neural networks, but rather to demonstrate that through our tensor sketching approach it is possible to design a substantially smaller network that achieves almost the same performance as the original network on a wide-range of datasets. We used the Torch machine learning framework and all the experiments were performed on a cluster of GPUs using a single GPU for each run. Additional experimental results are presented in Appendix~\ref{app:exp}. 

\noindent\textbf{Metrics.} We define {\em compression rate} as the ratio between the number of parameters in the reduced (compressed) network architecture and the number of parameters in the original (uncompressed) network architecture. Compression rate $< 1$ indicates compression with smaller values indicating higher compression.  The top-$1$ error (denoted by $\topone$) for a trained model on a test set captures the percentage of images in the test set misclassified by the model. To get a more stable picture of the model performance, $\topone$ is computed by averaging the test error after each of the last $10$ training epochs.

\noindent\textbf{Datasets.} We use 5 popular image datasets: CIFAR10  (objects recognition dataset with $3 \TIMES 32 \TIMES 32$ images),  SVHN (digits recognition dataset with $3 \TIMES 32 \TIMES 32$ images),  STL10 (objects recognition dataset with $3 \TIMES 96 \TIMES 96$ images), ImageNet10 objects recognition dataset with $3 \TIMES 256 \TIMES 256$ images, a subset of ImageNet1000 dataset that we created\footnote{We used following classes: bottle, cat, grasshopper, grocery, truck, chair, running shoes, boat, stove, and clock. The training set consists of 13000 images and the test set consists of 500 images}, and  Places2 (scene understanding dataset with $365$ classes and about $8$ million images in the training set). Note that, Places2 is a big and challenging dataset that was used in the recent ILSVRC 2016 ``Scene Classification'' challenge.

\noindent\textbf{Network Architectures.} We ran our experiments on four different network architectures. The choice of architectures was done keeping in mind limited computational resources at our disposal and a recent trend of moving away from fully connected layers in CNNs. A common observation in this area is that reducing the number of parameters in convolutional layers seems to be a much more challenging problem than that for fully connected layers. The first network architecture that we experiment with is the popular Network-in-Network (NinN)~\cite{nin} with minor adjustments for the corresponding image sizes (we used strides of the first layer to make these adjustments). Network-in-Network is a moderately sized network which attains good performance on medium sized datasets, e.g. CIFAR10~\cite{cifar10nin}. For this network, we did not employ batch normalization~\cite{ioffe2015batch} or dropout~\cite{srivastava2014dropout} to have a uniform set of experiments across different techniques. The second network that we consider is the same as NinN with only one change that the last convolution layer is replaced by a fully connected layer (we denote it as NinN+FC). Following~\cite{HashedNets}, the third network that we experiment is a simple shallow network, which we refer to as TestNet, with only 2 convolution layers and 2 fully connected layers which allows us to easily test the efficacy of our approximation technique for each layer individually. We describe the construction of TestNet in more detail in Appendix~\ref{app:exp}. Table~\ref{t:base} shows the original (uncompressed) top-$1$ error ($\topone$) for NinN and NinN+FC. The number of parameters are about 966K for NinN and 1563K for NinN+FC for all datasets. The statistics about TestNet are presented in Figure~\ref{t:basetestnet} (Appendix~\ref{app:exp}). The final network that we consider is GoogLeNet~\cite{szegedy2015going} with batch normalization, which we use for the Places2 dataset. This network has a top-$1$ error of 32.3\% on the Places2 dataset.
\begin{table}[ht]
\begin{center}
\small
\begin{tabular}{c|cccc}
\hline
Network & CIFAR10 &STL10   & SVHN  & ImageNet10 \\
\hline
NinN  & 17.7 &  43.2  &   6.0  & 27.1 \\
NinN+FC & 16.9 &  41.2 &    5.4  & 26.0 \\
\hline
\end{tabular}
\vspace*{-1ex}
\caption{Top-$1$ error of the NinN architecture and its variant on different datasets.} 
\label{t:base}
\end{center}
\vspace*{-1ex}
\end{table}

\noindent\textbf{Baseline Techniques.} As discussed in Section~\ref{sec:related} there are by now quite a few techniques for network approximation. We compare our proposed approach with four state-of-the-art techniques that approximate {\em both} the convolutional and the fully connected layers: FreshNets technique  that uses hashing in the frequency domain to approximate the convolutional layer~\cite{FreshNets}, low-rank decomposition technique of~\cite{denil2013predicting} ($\LOWRANK_1$), and tensor decomposition technique of~\cite{tai2015convolutional} ($\LOWRANK_2$). While using the FreshNets technique, we also use the HashedNets technique of feature hashing~\cite{HashedNets} for compressing the fully connected layers as suggested by~\cite{FreshNets}. We used open-source implementations of all these techniques: HashedNets, FreshNets, and $\LOWRANK_1$ are from~\cite{FN} and $\LOWRANK_2$ from~\cite{tai}. We set the required parameters to ensure that all the compared approaches achieve about the same compression rate. 

\begin{figure*}[ht]
\hspace*{-.75in}
\begin{tabular}{cccc}
\begin{subfigure}{0.28\textwidth}\centering\includegraphics[width=1.1\columnwidth]{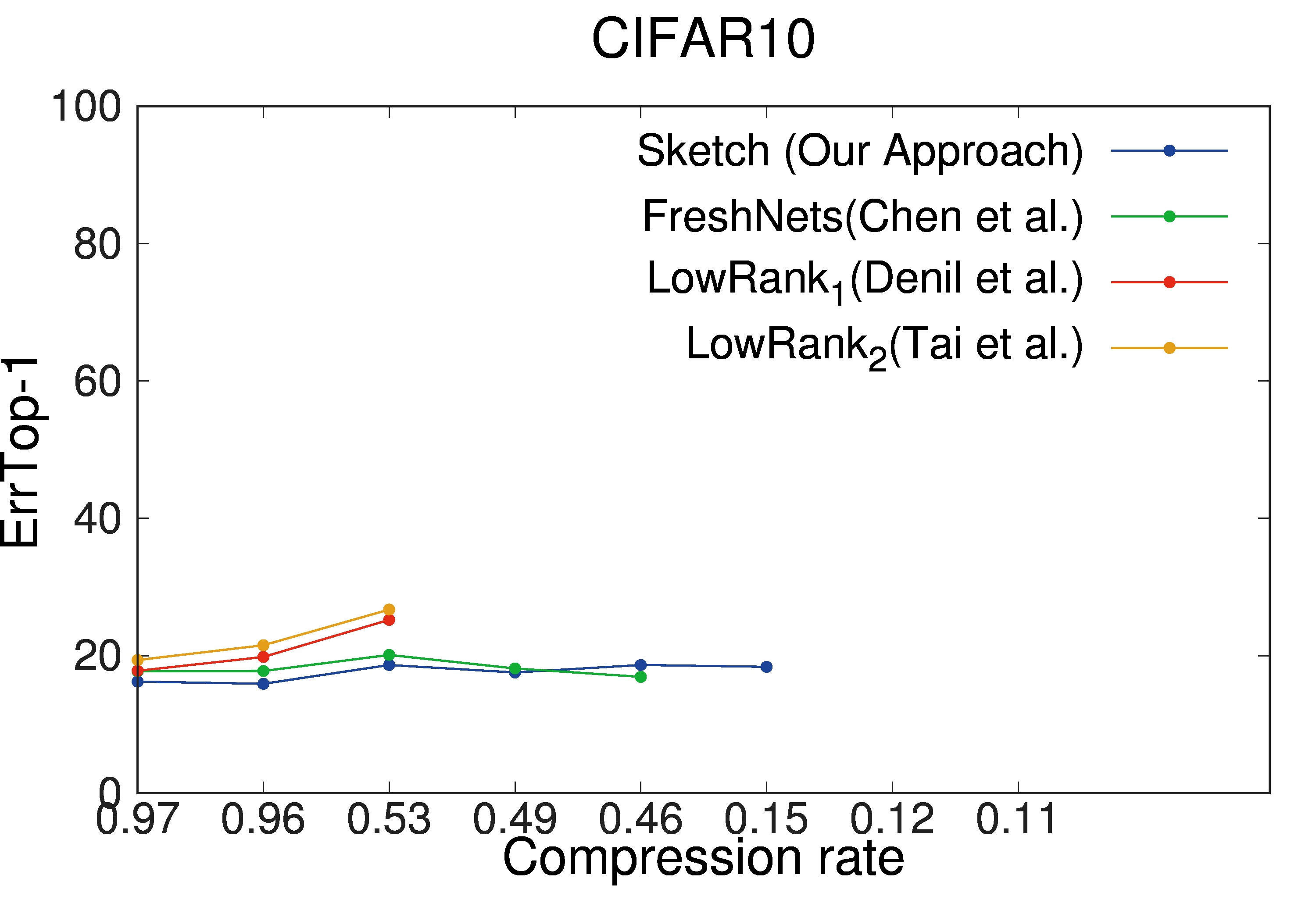}\end{subfigure}&
\begin{subfigure}{0.28\textwidth}\centering\includegraphics[width=1.1\columnwidth]{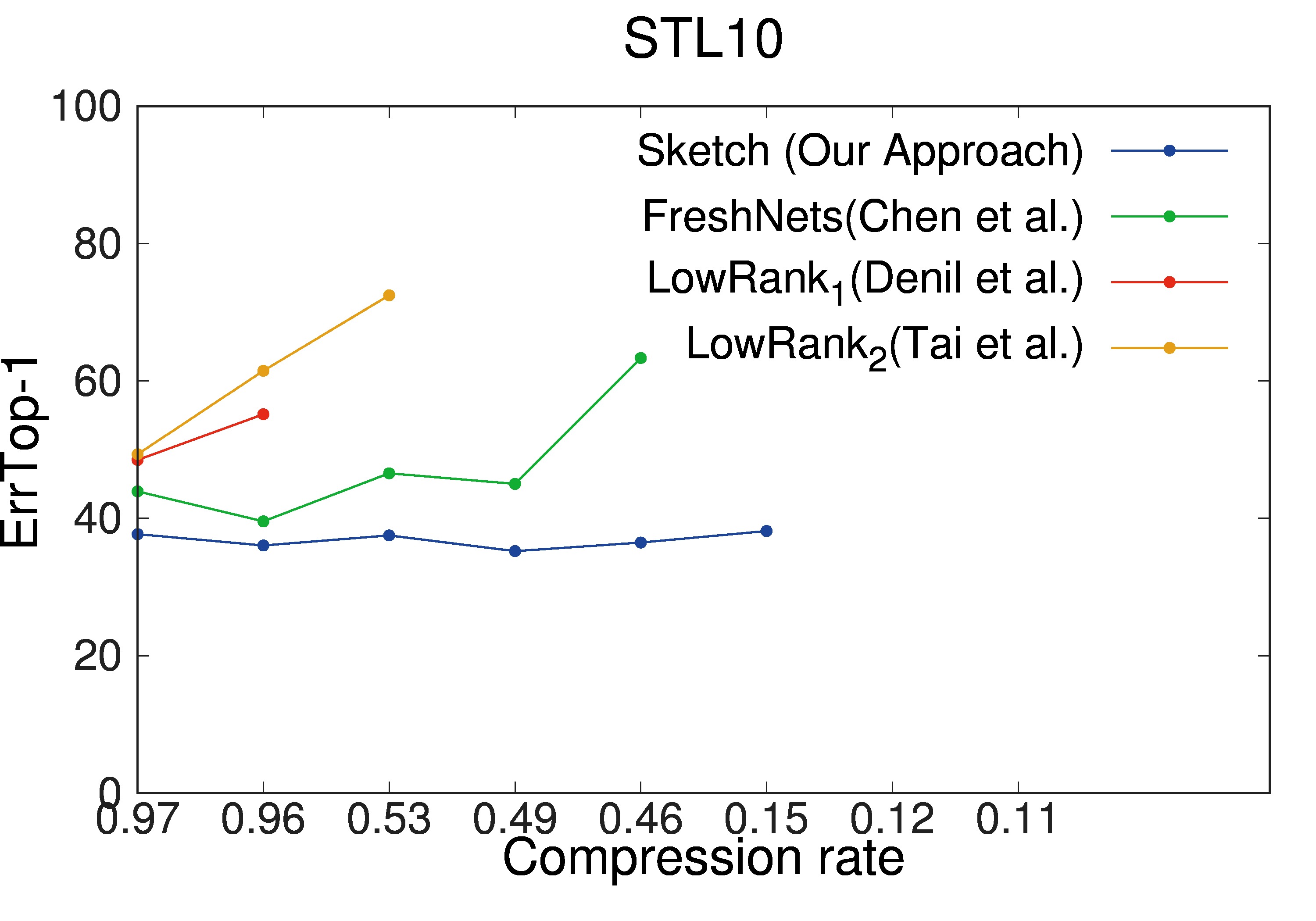}\end{subfigure} &
\begin{subfigure}{0.28\textwidth}\centering\includegraphics[width=1.1\columnwidth]{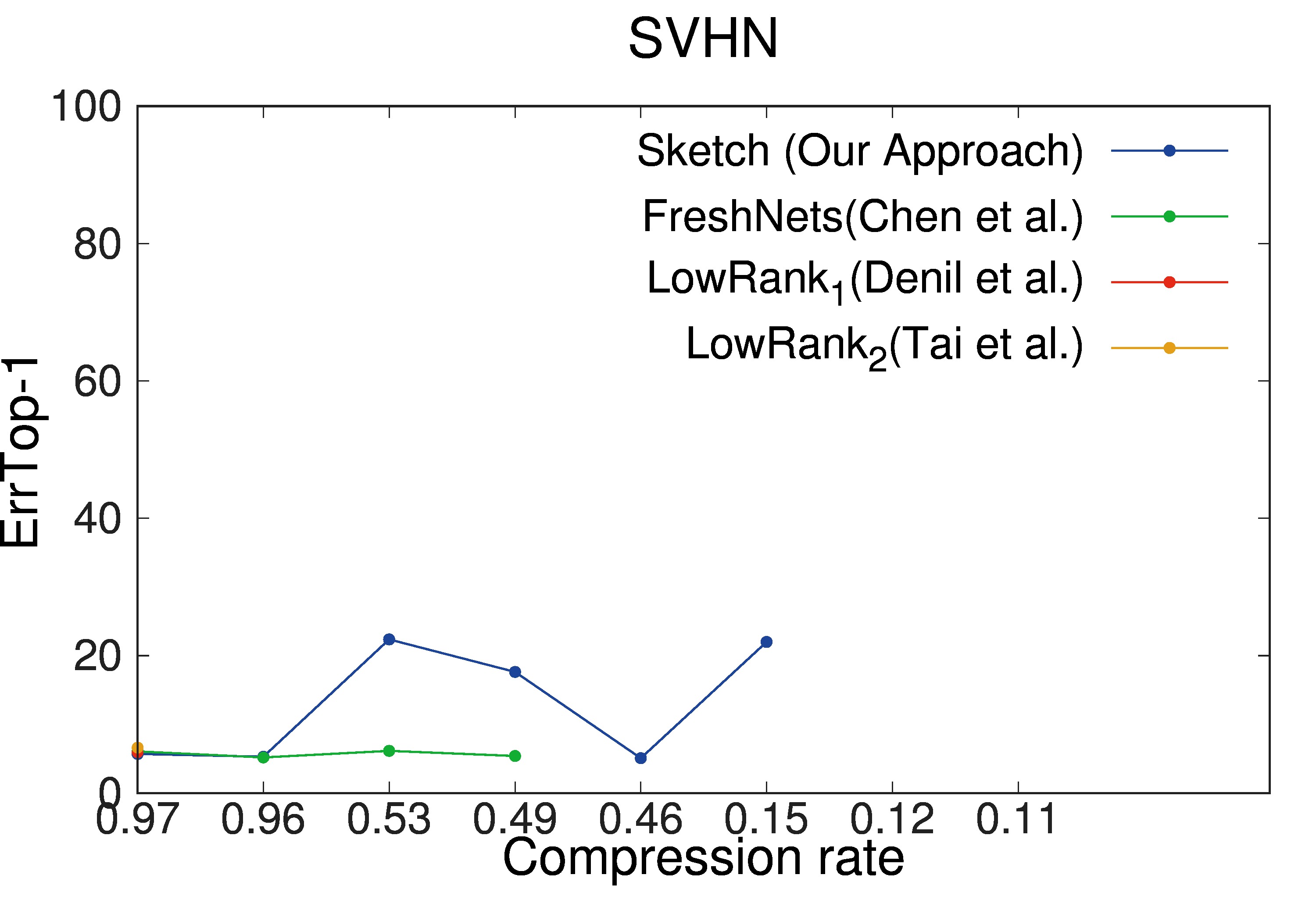}\end{subfigure}&
\begin{subfigure}{0.28\textwidth}\centering\includegraphics[width=1.1\columnwidth]{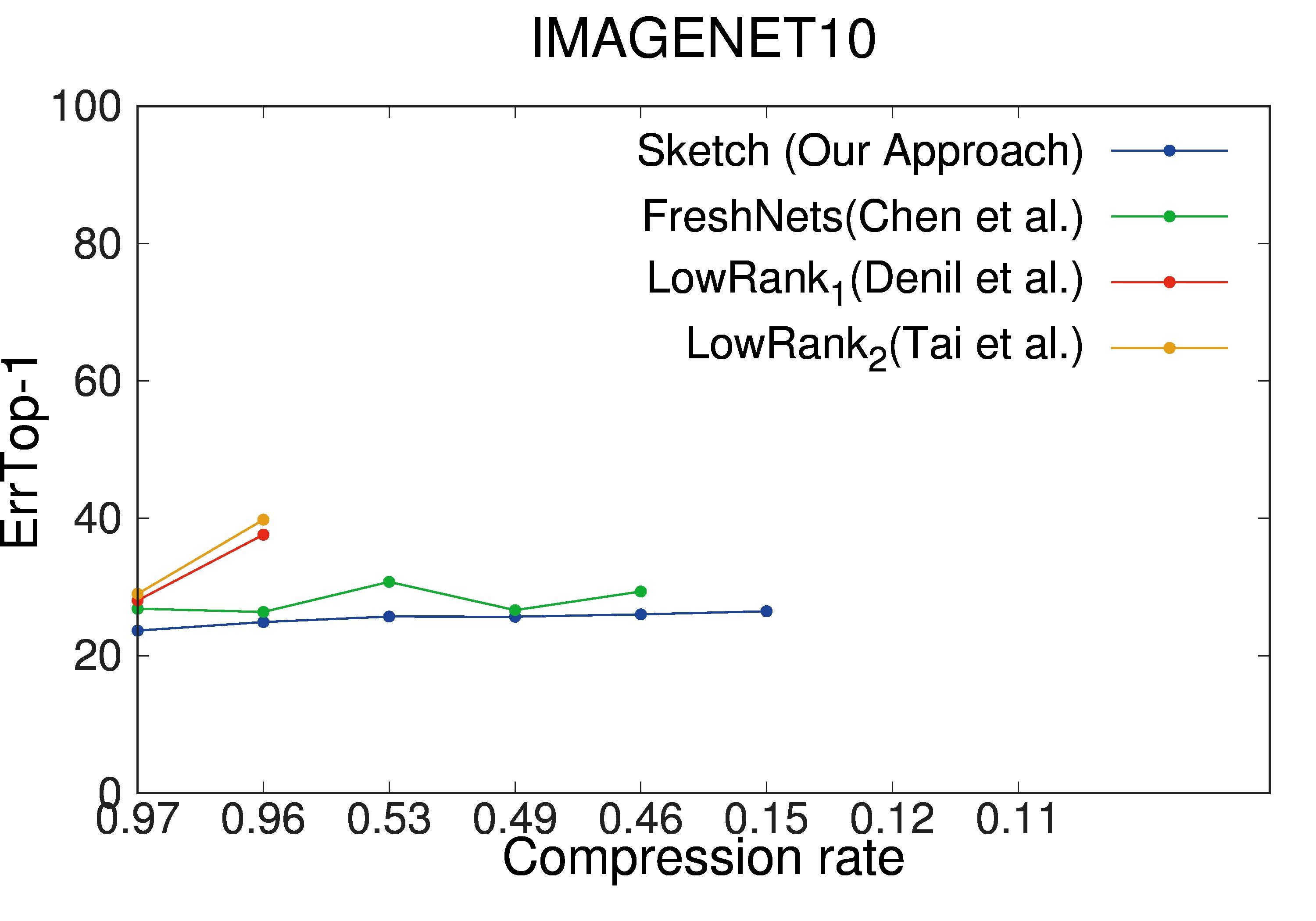}\end{subfigure}
\end{tabular}
\vspace*{-1ex}
\caption{Top-$1$ error for the NinN architecture as we decrease the compression rate by compressing one convolutional layer at a time each by a factor of $10$. The x-axis is not to scale.}
\label{t:conv}
\hspace*{-.75in}
\begin{tabular}{cccc}
\begin{subfigure}{0.28\textwidth}\centering\includegraphics[width=1.1\columnwidth]{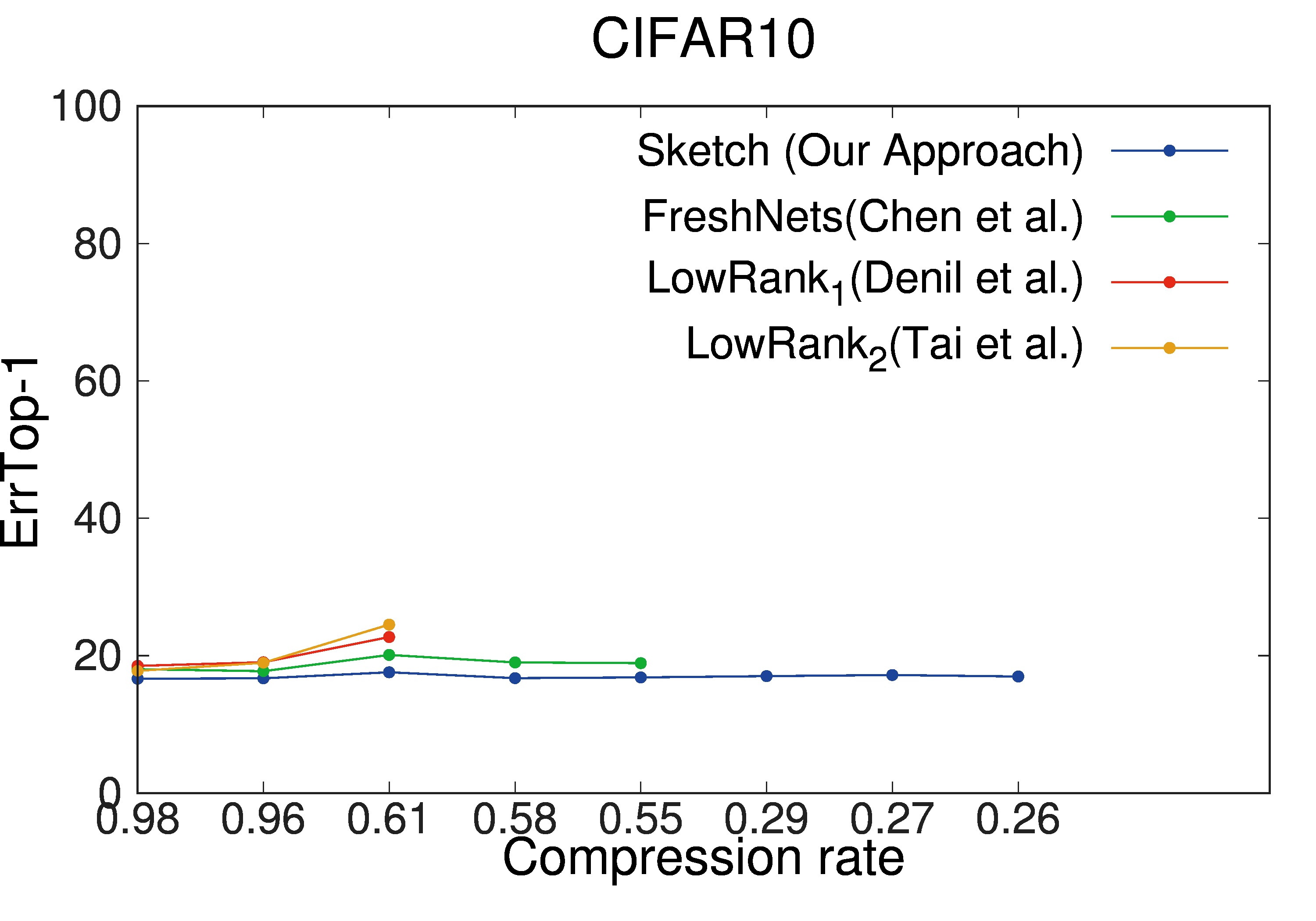}\end{subfigure}&
\begin{subfigure}{0.28\textwidth}\centering\includegraphics[width=1.1\columnwidth]{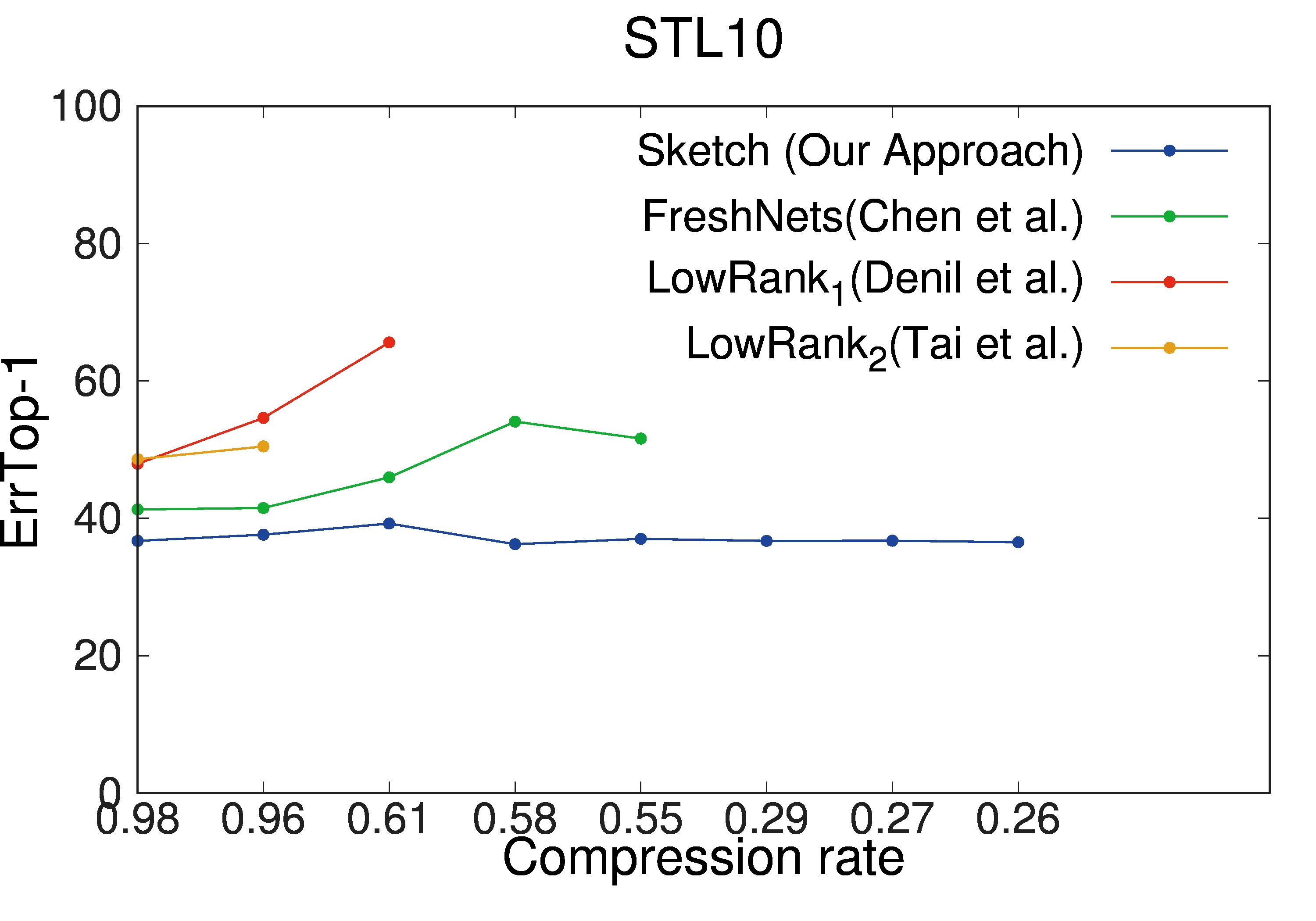}\end{subfigure}&
\begin{subfigure}{0.28\textwidth}\centering\includegraphics[width=1.1\columnwidth]{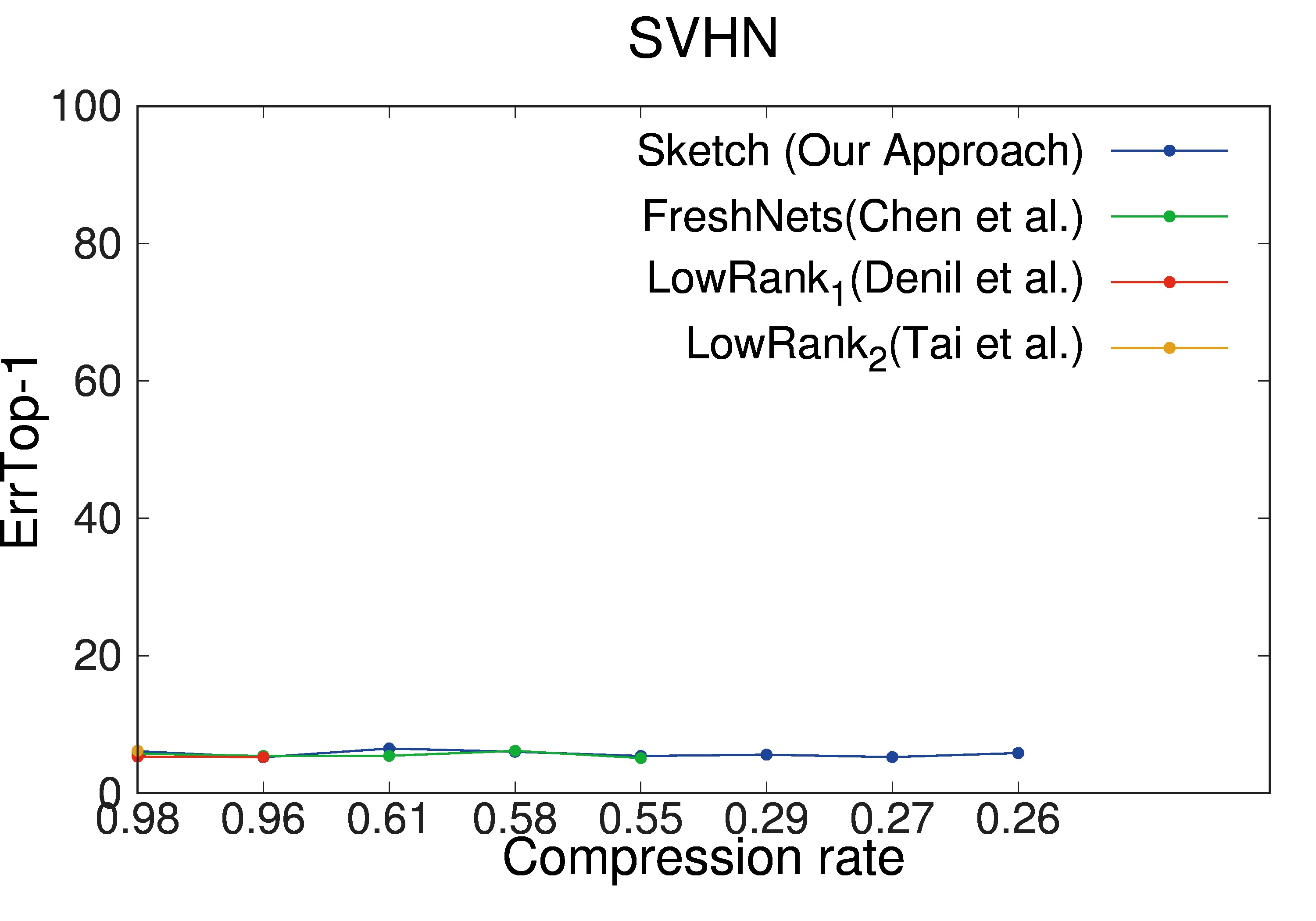}\end{subfigure}& 
\begin{subfigure}{0.28\textwidth}\centering\includegraphics[width=1.1\columnwidth]{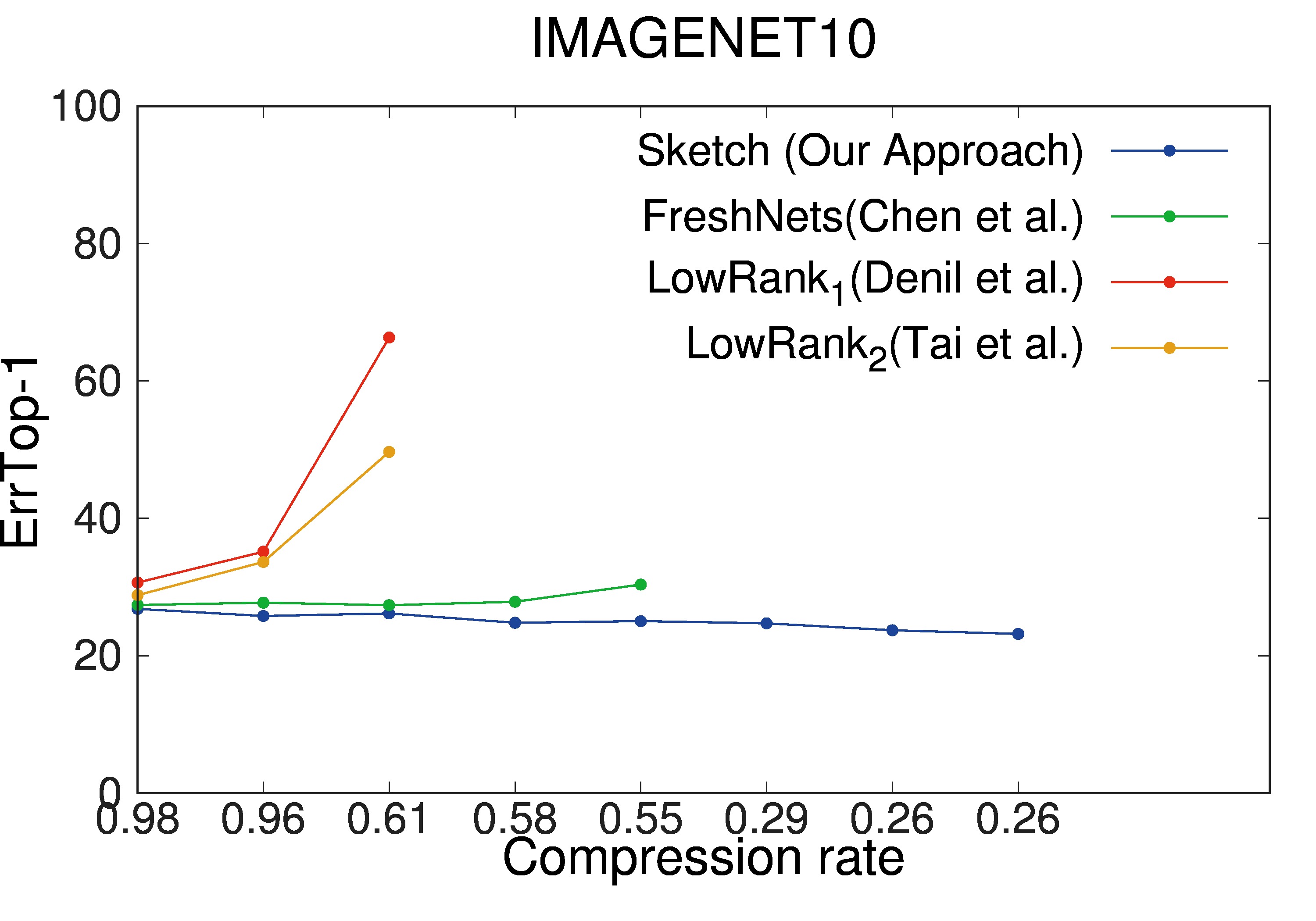}\end{subfigure}
\end{tabular}
\vspace*{-1ex}
\caption{Top-$1$ error for the NinN architecture as we decrease the compression rate by compressing one convolutional layer at a time each by a factor of~$4$. The x-axis is not to scale.}
\label{t:conva}
\hspace*{-.75in}
\begin{tabular}{cccc}
\begin{subfigure}{0.28\textwidth}\centering\includegraphics[width=1.1\columnwidth]{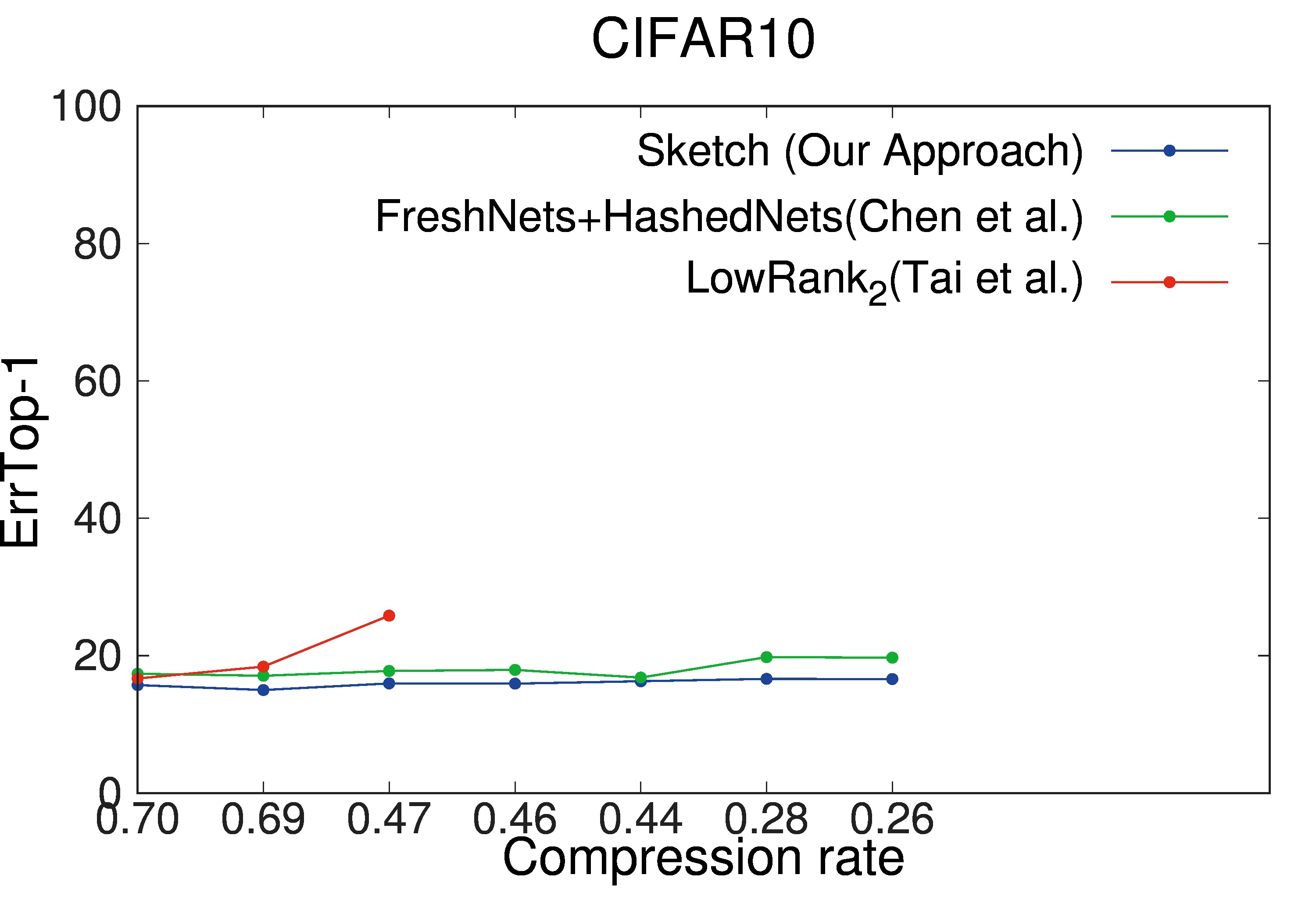}\end{subfigure}&
\begin{subfigure}{0.28\textwidth}\centering\includegraphics[width=1.1\columnwidth]{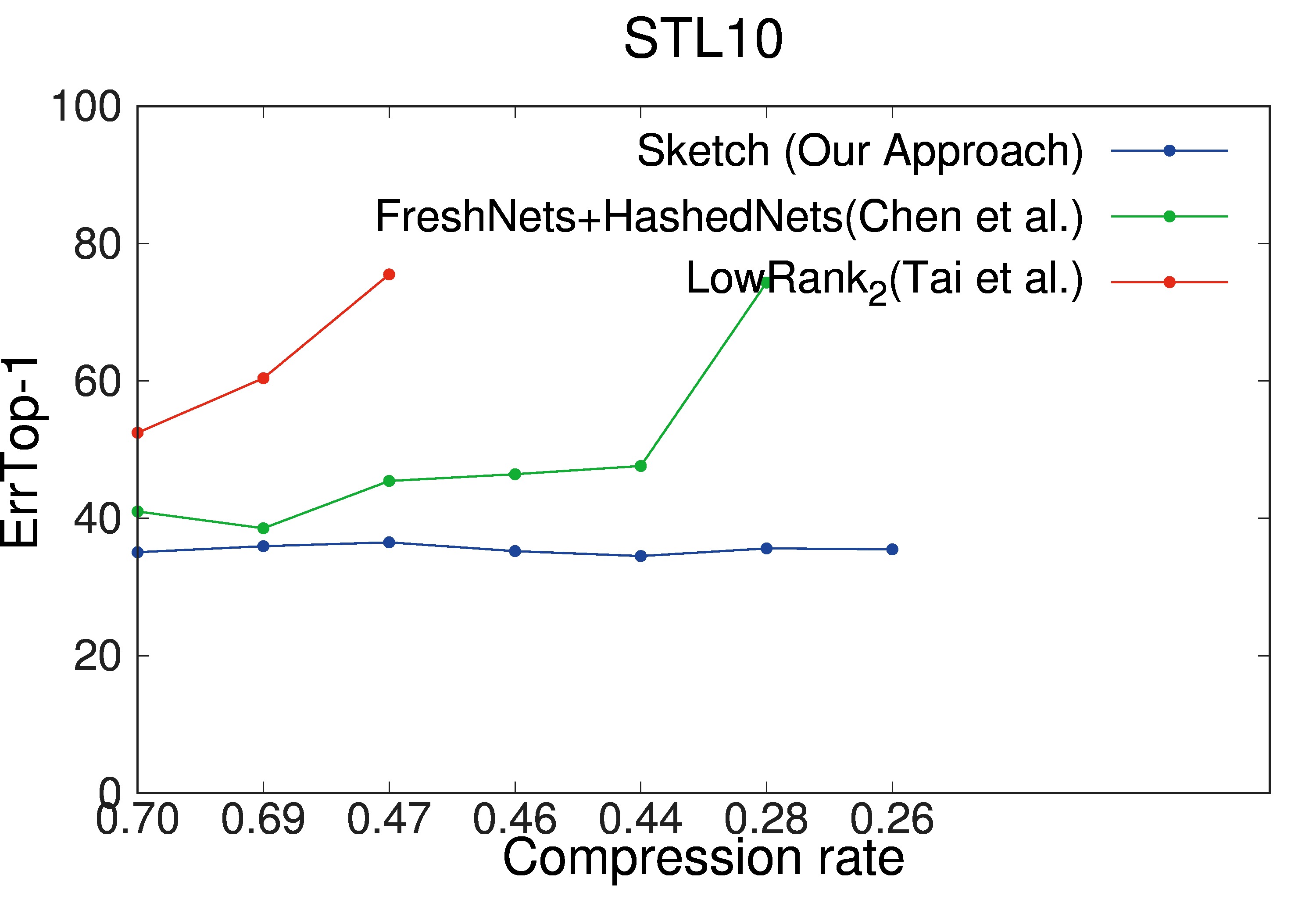}\end{subfigure}&
\begin{subfigure}{0.28\textwidth}\centering\includegraphics[width=1.1\columnwidth]{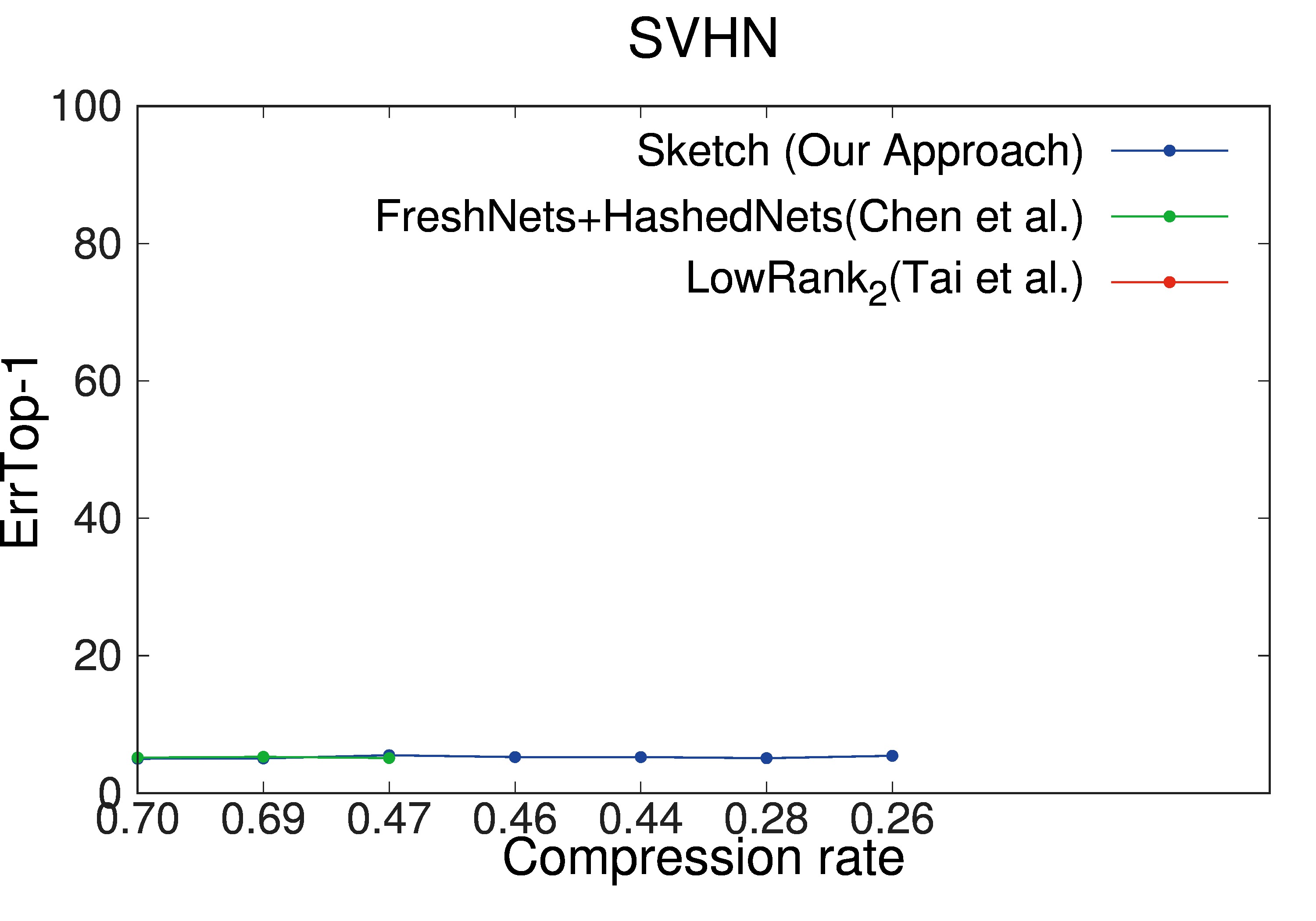}\end{subfigure}&
\begin{subfigure}{0.28\textwidth}\centering\includegraphics[width=1.1\columnwidth]{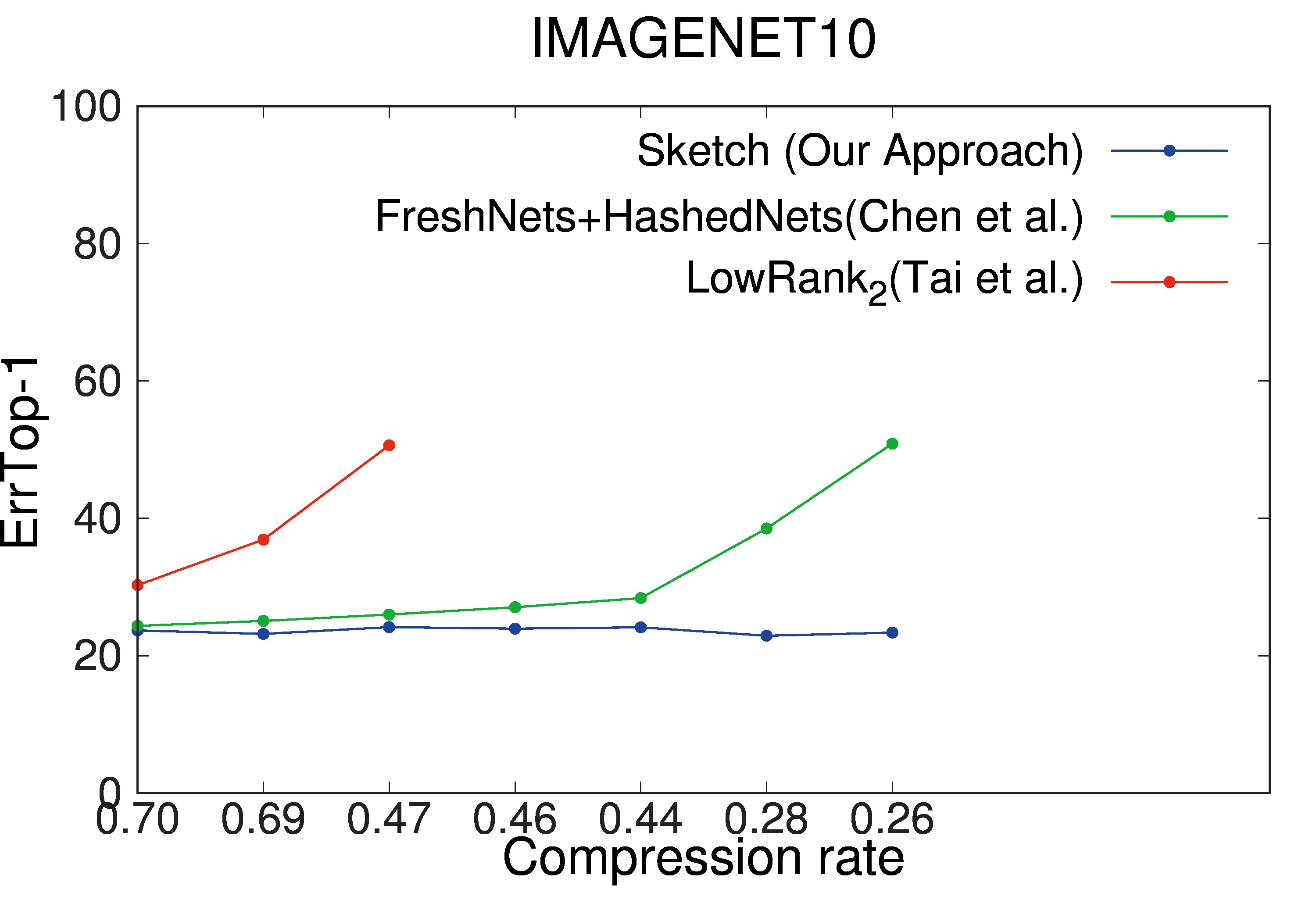}\end{subfigure}
\end{tabular}
\vspace*{-1ex}
\caption{Top-$1$ error for the NinN+FC architecture. The size of $\FC$ layer is about half of the total size of convolutional layers $\Conv_2$ to $\Conv_8$. We compress the fully connected layer by a factor of $4$. We then use a similar experimental setup as in Figure~\ref{t:conva} of reducing the number of parameters in the convolutional layers ($\Conv_2$ to $\Conv_8$) each by a factor of $4$. The x-axis is not to scale.}
\label{t:fcconv}
\end{figure*}

\noindent\textbf{Compression on the Convolutional Layers.} We performed a set of experiments to evaluate the performance of our scheme only on the convolutional layers. We used the NinN architecture for this purpose. NinN is, essentially, a sequence of  nine convolution layers (labeled as $\Conv_1$ to $\Conv_9$). We compress these layers one by one, starting from $\Conv_2$ and finishing at $\Conv_9$ by reducing the number of parameters in each layer by a factor of $r$ which is set as $10$. When all these $8$ convolution layers are compressed the achieved network compression rate is approximately\footnote{We do not compress the first layer that takes input.} equal to $1/r$.

Figures~\ref{t:conv} and~\ref{t:conva} shows the results of our experiments. If a point is missing in the plots then the corresponding network training failed. We expect the error to go up as we decrease the compression rate, i.e., increase the parameter reduction. We observe this general trend in almost all our plots, with minor fluctuations such as in Figure~\ref{t:conv} on the SVHN dataset.
We make two main observations from these plots. First, our method was always able to get to a better compression rate compared to other techniques, in that these comparative techniques started failing sooner as we kept decreasing the compression rate. For example, our approach consistently achieves a compression rate of $0.15$ that none of the other techniques even get close to achieving. Second, our approach also almost always achieves better accuracy when compared to other techniques.  As explained in Section~\ref{sec:related}, our approach has some advantages over the compared techniques, especially  in terms of its ability to approximate (compress) the convolutional layers.
Effects of this become more pronounced as we decrease the compression rate. In most cases, we gain up to $4\%$ or lose up to $2\%$ of accuracy compared to original network accuracy. The fact that sometimes our reduced network was able to gain a bit of accuracy over the original network suggests that our randomized technique probably also adds a regularization effect during the training. 

\noindent\textbf{Compression on both the Convolutional and Fully Connected Layers.} We now add fully connected layers into the mix. To do so, we used a modified NinN architecture (denoted as NinN+FC) in our experiments where we replaced the last convolution layer ($\Conv_9$) with a fully connected layer of size $768 \times 768$ followed by a classifier layer of size $768 \times 10$. In Figure~\ref{t:fcconv}, we present the results of these experiments. Our approach again outperforms other techniques in terms of both accuracy and the maximum achievable compression rate. The results demonstrate the effectiveness of proposed approach on both the convolutional and fully connected layers.

\noindent\textbf{Places2 Dataset.} To evaluate our approach on a large dataset, we ran additional experiments on the Places2 dataset (using a centered crop). Here we used the GoogLeNet architecture with batch normalization. Due to limited computational resources, we ran a single experiment where we compressed all but the first layer to achieve a compression rate of about $0.2$. At this compression level, training for none of the competitor methods succeeded, whereas, our approach gave a top-$1$ error of 36.4\%. Note that the top-$1$ error of the original GoogLeNet on this dataset is 32.3\%. This demonstrates that our approach manages to generate smaller networks that perform well even on large datasets. Again here, as in all the above cases, model storage sizes can be further reduced by taking this reduced model and using certain post-processing operations as detailed in Section~\ref{sec:related}, which is outside the scope of this evaluation.

\noindent\textbf{Parameter Sensitivity.} In Appendix~\ref{app:exp}, we present experiments that highlight the role of parameters $k$ and $\ell$ in our proposed approach. In general, we observe that the accuracy of the compressed models improve as we increase $k$ or $\ell$ (this happens because we are increasing the effective size of the constructed sketches). Also, due to the averaging effect, increasing $\ell$ decreases the variance of top-$1$ error with respect to the randomization that arises from the use of random matrices.

\noindent\textbf{Computational Efficiency.} While our primary focus is on  network approximation (i.e., designing networks with a smaller set of parameters), an added bonus is that the networks generated through our tensor sketching approach are also computationally more efficient. For example, at the compression rate of $0.15$ the wall-clock testing time, of our reduced NinN is on average between $1.6$-$2$x smaller compared to the original network across all the tested datasets. Since the sketch tensors in our construction are dense, further efficiency gains are possible by better exploiting the dense matrix capabilities of modern GPUs.


\appendix
\section{Proof of Proposition~\ref{prop:prop}} \label{app:tensor}

\begin{proposition} [Proposition~\ref{prop:prop} Restated]
Let $W \in \R^{d_1 \times d_2}$. Let $U_1 \in \R^{k \times d_1}$ and $U_2 \in \R^{k \times d_2}$ be two independent random scaled sign matrices. Let $S_1 = U_1 W (= W \times_1 U_1) $ and $S_2 =W U_2^\top (= W \times_2 U_2)$. Then for any matrix $M \in \R^{d_2 \times d_3}$:
\begin{CompactEnumerate}
\item \label{part1} $\E[U_1^\top S_1 M ] = W M \mbox{ and } \;\; \E[S_2 U_2 M] = W M.$
\item \label{part2}  $\E \left [ \left \| U_1^\top S_1 M - W M  \right \|_F^2 \right] \leq \frac{2 d_1\|WM\|_F^2 }{k}, \mbox{ and}$
\item[] $\;\;\; \E \left [ \left \| S_2 U_2 M - W M \right \|_F^2 \right] \leq \frac{2\|W\|_F^2 \| M \|_F^2}{k}.$
\end{CompactEnumerate}
\end{proposition}
\begin{proof}
Part~\ref{part1} follows by simply using linearity of expectation.

We focus on Part~\ref{part2} which investigates the variance bounds for $U_1^\top S_1 M$ and $S_2 U_2 M$. For this, we use some standard ideas from the matrix sketching literature~\cite{TCS-060}.

Consider first $\E \left [\| S_2 U_2 M - W M \|_F^2 \right]$. We start by noting,
\begin{align*}
S_2 U_2 M - W M  & = W U_2^\top U_2 M - WM  = \frac{1}{k}W Z ^\top Z M - WM,
\end{align*}
where $U_2 = Z/\sqrt{k}$. Let $\w_a,\z_b, \m_c$ denote the $a, b, c$-th columns of $W^\top, Z$, and $M$ respectively. We have 
$$\left \| W Z ^\top Z M  \right \|_F^2 = \sum_{a,c} \left (\w_a^\top  \left( \sum_{b=1}^k \z_b \z_b^\top \right ) \m_c \right )^2.$$ 
Therefore, we get
\begin{align} \label{eqn:frob}
\left \| \frac{1}{k}W Z ^\top Z M - WM  \right \|_F^2 & = \sum_{a,c} \left (\frac{1}{k} \w_a^\top  \left( \sum_{b=1}^k \z_b \z_b^\top \right ) \m_c -  \w_a^\top \m_c \right )^2 \nonumber \\
& = \sum_{a,c} \left (\frac{1}{k} \w_a^\top  \left( \sum_{b=1}^k \z_b \z_b^\top \right ) \m_c - \frac{1}{k} \sum_{b=1}^k \w_a^\top \m_c \right )^2 \nonumber\\
& = \sum_{a,c} \left ( \sum_{b=1}^k \frac{\w_a^\top \z_b \z_b^\top \m_c - \w_a^\top \m_c}{k}  \right )^2.
\end{align} 
Let $y_{abc} = \frac{\w_a^\top \z_b \z_b^\top \m_c - \w_a^\top \m_c}{k}$ which can be re-expressed as: 
$$y_{abc}= \frac{1}{k} \sum_{\substack{r,s \\ r \neq s}} W_{ar} Z_{rb} Z_{sb} M_{sc},$$
where $W_{ar}$ is the $(a,r)$th entry in $W$, $Z_{rb}$ and $Z_{sb}$ are the $(r,b)$th and $(s,b)$th entries in $Z$ respectively, and $M_{sc}$ is the $(s,c)$th entry in $M$. Using this notation, we can re-express~\eqref{eqn:frob} as:
\begin{align*}
 \left \| \frac{1}{k}W Z ^\top Z M - WM \right \|_F^2 & = \sum_{a,c} \left (\sum_{b=1}^k y_{abc} \right )^2 \\
& = \sum_{a,c} \sum_{b,b'} y_{abc} \, y_{ab'c} \\
& = \frac{1}{k^2} \sum_{a,c} \sum_{b,b'} \sum_{r \neq s} W_{ar} Z_{rb} Z_{sb} M_{sc}   \sum_{r' \neq s'} W_{ar'} Z_{r'b'} Z_{s'b'} M_{s'c} . 
\end{align*}
Taking expectation, 
\begin{align*}
\E\left[  \| S_2 U_2 M - W M \|_F^2 \right] = \frac{1}{k^2} \sum_{\substack{a,c,b,b',r,s,r',s' \\ r \neq s \\ r' \neq s'}}  W_{ar} W_{ar'} M_{sc} M_{s'c} \, \E[Z_{rb}  Z_{sb} Z_{r'b'}  Z_{s'b'}].
\end{align*}
Now, $\E[Z_{rb}  Z_{sb} Z_{r'b'}  Z_{s'b'}]$ is non-zero only if either: 1) $r=r'$, $s=s'$, and $b=b'$ or 2) $r=s'$, $s=r'$, and $b=b'$. Therefore, we can simplify $\E\left [\| S_2 U_2 M - W M \|_F^2 \right]$ as,
\begin{align} \label{eqn:firstprev} 
\E \left [ \left \| S_2 U_2 M - W M  \right \|_F^2 \right ] & \leq  \frac{2}{k^2} \sum_{a,c} \sum_{b} \sum_{r} W_{ar}^2 \sum_{s} M_{sc}^2 \nonumber \\ 
& = \frac{2}{k} \sum_{a,r} W_{ar}^2 \sum_{c,s} M_{sc}^2 \nonumber \\
&= \frac{2}{k} \| W \|_F^2 \| M \|_F^2.
\end{align}
This proves the claimed bound for $\E \left [\| S_2 U_2 M - W M \|_F^2 \right]$. 

Now we bound $\E \left [\| U_1^\top S_1 M - W M \|_F^2 \right]$. We start by re-expressing the result in~\eqref{eqn:firstprev}. Start by noting that $S_2 = W U_2^\top$. Therefore, from~\eqref{eqn:firstprev},
\begin{align*} 
\E \left [\left \| W U_2^\top U_2 M - W M \right \|_F^2 \right] \leq \frac{2}{k} \| W \|_F^2 \| M \|_F^2.
\end{align*}
Now by setting, $W = \mathbb{I}_{d_1}$ in this result and by noting $\| \mathbb{I}_{d_1} \|_F^2 = d_1$, we get that for any matrix $A \in \R^{d_2 \times d_3}$,
\begin{align} \label{eqn:prev} 
\E \left [ \left \| U_1^\top U_1 A - A \right \|_F^2 \right] \leq \frac{2 d_1}{k} \| A \|_F^2,
\end{align}
where the expectation is now over $U_1 \in \R^{k \times d_1}$.

Since $U_1^\top S_1 M = U_1^\top U_1 W M$. Therefore, $U_1^\top S_1 M - W M = U_1^\top U_1 W M -  W M$. The idea is to invoke~\eqref{eqn:prev} with $A = WM$. We get,
$$ \E \left [ \left \| U_1^\top U_1 W M - WM \right \|_F^2 \right] \leq \frac{2 d_1}{k} \| W M \|_F^2.$$
This completes the proof of this theorem.
\end{proof}

\section{Proof of Theorem~\ref{thm:convprop}} \label{app:convprop}
\begin{theorem} [Theorem~\ref{thm:convprop} Restated]
Let $\mathcal{K} \in \R^{d_2 \times h \times w \times d_1}$ be a kernel tensor and $K = \mat_4(\mathcal{K})$. Let $U_1 \in \R^{k \times d_1}$ and $U_2 \in \R^{k h w \times d_2 h w}$ be two independent random scaled sign matrices. Let $\mathcal{S}_1$ and $\mathcal{S}_2$ be tensors such that $\mat_4(\mathcal{S}_1) = K \times_2 U_1$ and $\mat_4(\mathcal{S}_2) = K \times_1 U_2$. Then for any input matrix $I_\inp \in \R^{h_2 w_2 \times d_2 h w}$ (formed from an input tensor $\mathcal{I}_\inp \in \R^{h_1 \times w_1 \times d_2}$):
\begin{CompactEnumerate}
\item Unbiased Estimation: $\E[I_\inp\, \mat_4(\mathcal{S}_1) U_1] = I_\inp K$, and  $\E[I_\inp U_2^\top \, \mat_4(\mathcal{S}_2)] = I_\inp K.$
\item Variance Bound:
\vspace*{-1ex} 
\begin{align*}
& \E  \left [ \left \| I_\inp \, \mat_4(\mathcal{S}_1) U_1 - I_\inp K \right \|_F^2 \right ] \leq \frac{2 d_1\|I_\inp K\|_F^2 }{k}, \mbox{ and} \\
&\E \left [ \left \| I_\inp U_2^\top \, \mat_4(\mathcal{S}_2) - I_\inp K \right \|_F^2 \right ] \leq \frac{2 \|I_\inp\|_F^2 \| K \|_F^2}{k h w}.
\end{align*}
\end{CompactEnumerate}
\end{theorem}
\begin{proof}
First note that, by definition,
\begin{align*}
I_\inp\, \mat_4(\mathcal{S}_1) U_1 = I_\inp K U_1^\top U_1.
\end{align*} 
Using an analysis similar to Proposition~\ref{prop:prop} gives,
\begin{eqnarray*}
& \E[I_\inp\, \mat_4(\mathcal{S}_1) U_1] = I_\inp K, \mbox{ and } & \\
& \E \left [ \left \| I_\inp\, \mat_4(\mathcal{S}_1) U_1 - I_\inp K \right \|_F^2 \right] \leq \frac{2 d_1\|I_\inp K\|_F^2 }{k}.& 
\end{eqnarray*}
Similarly, by definition of $\mat_4(\mathcal{S}_2)$, we have:
\begin{align*}
I_\inp U_2^\top \, \mat_4(\mathcal{S}_2) = I_{\inp} U_2^\top U_2 K.
\end{align*}
Again relying on an analysis similar to Proposition~\ref{prop:prop} gives,
\begin{eqnarray*}
& \E[I_\inp U_2^\top \, \mat_4(\mathcal{S}_2)] = I_\inp K, \mbox{ and } & \\
& \E \left [\left \| I_\inp U_2^\top \, \mat_4(\mathcal{S}_2) - I_\inp K \right \|_F^2 \right ] \leq \frac{2\|I_\inp\|_F^2 \| K \|_F^2}{k h w}.& 
\end{eqnarray*}
This completes the proof of this theorem.
\end{proof}

\section{Additional Experimental Results} \label{app:exp}
In this section, we present some additional experimental results that investigate the role of various parameters in our proposed approach. We start by describing the TestNet architecture that we use for the following experiments. 

\begin{figure*}[!ht]
\hspace*{-.2in}
\scriptsize
\begin{tabular}{ccccccc}
\hline
Images&$\Conv_1$ &  $\MAXPOOL_1$ & $\Conv_2$ &  $\MAXPOOL_2$ & $\FC_1$ &  $\FC_2$ \\
\hline
$3 \TIMES 32 \TIMES 32$ & $d_2=3, d_1=30, f=5\TIMES5$ & $f=2\TIMES2$ & $d_2=30, d_1=30, f=5\TIMES5$ & $f=4\TIMES4$ & $d_2=480, d_1=250$ & $d_2=250, d_1=10$\\
$3\TIMES 96 \TIMES 96$ & $d_2=3, d_1=20, f=7\TIMES7$ & $f=5\TIMES5$ & $d_2=20, d_1=40, f=5\TIMES5$ & $f=5\TIMES5$ & $d_2=1960, d_1=500$ & $d_2=500, d_1=10$\\
$3\TIMES 256 \TIMES 256$ & $d_2=3, d_1=20, f=7\TIMES7$ & $f=11\TIMES11$ & $d_2=20, d_1=30, f=9\TIMES9$ & $f=7\TIMES7$ & $d_2=3000, d_1=500$ & $d_2=500, d_1=10$ \\
\hline
\end{tabular}
\caption{TestNet Architecture.} 
\label{t:testnetparams}
\end{figure*}

\paragraph{TestNet Architecture.} TestNet is a simple shallow network with only 2 convolution layers and 2 fully connected layers. This allows us to easily test the efficacy of our approximation technique for each layer individually. Figure~\ref{t:testnetparams} shows parameters of TestNet for different image sizes. 

A ReLU layer is used after each fully connected and convolutional layer. For example, consider images of size $3 \TIMES 32 \TIMES 32$. The first convolutional layer takes $3$ input feature maps ($d_2=3$) and produces $30$ output feature maps ($d_1=30$) using filters of size $5$ by $5$ ($f = 5\TIMES5$), and we represent it as a $4$-dimensional tensor in $\R^{3 \times 5 \times 5 \times 30}$. Note that in TestNet the fully connected layers contain much more network parameters than the convolutional layers. 

Table~\ref{t:basetestnet} shows the original  top-$1$ error ($\topone$) and the number of parameters for all datasets. We used different number of parameters in $\FC_1$ for different image sizes to ensure that the corresponding trained networks converge.
\begin{table}[!h]
\begin{center}
\small
\begin{tabular}{cccc}
\hline
 CIFAR10 &STL10   & SVHN  & ImageNet10\\ \hline
$25.7$ (147K)& $40.5$ (1008K)  &  $8.2$  (147K) & $27.0$  (1561K) \\
\hline
\end{tabular}
\caption{Top-$1$ error of the original TestNet on different datasets. In bracket, we show the number of parameters in each of these networks.
} 
\label{t:basetestnet}
\end{center}
\end{table}

\begin{figure*}[!ht]
\begin{center}
\begin{tabular}{cc}
\begin{subfigure}{0.43\textwidth}\centering\includegraphics[width=0.95\columnwidth]{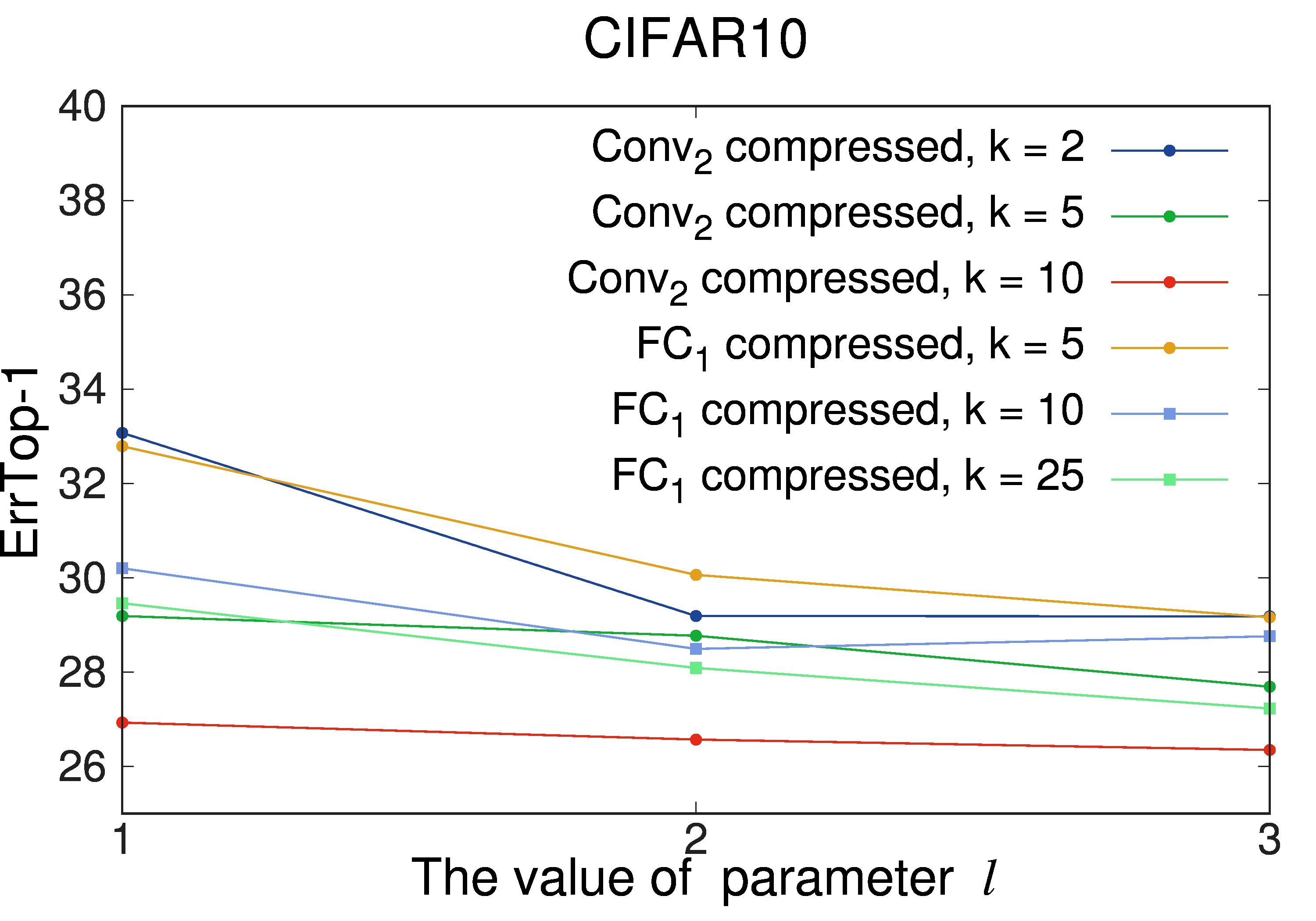}\end{subfigure}&
\begin{subfigure}{0.43\textwidth}\centering\includegraphics[width=0.95\columnwidth]{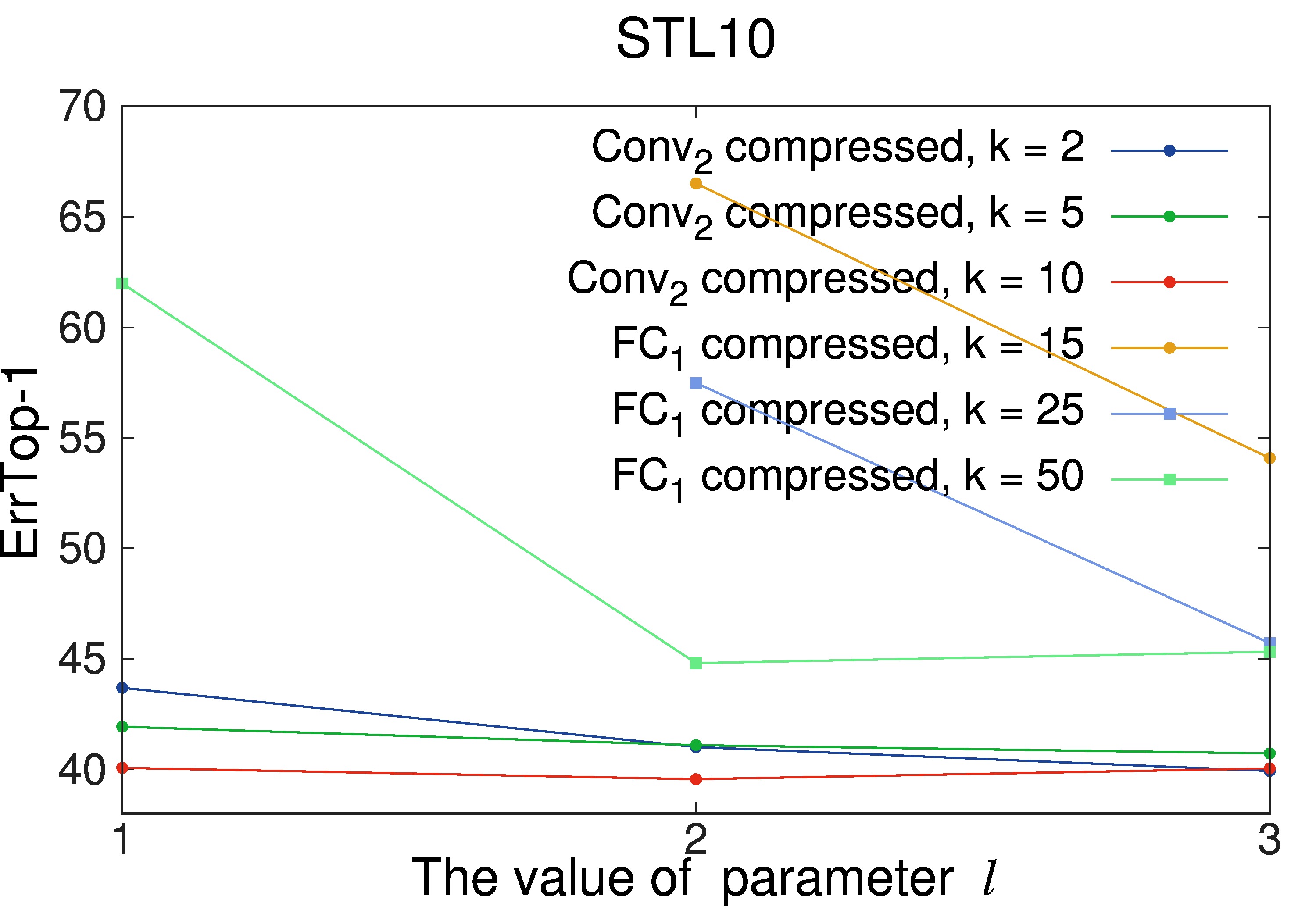}\end{subfigure}\\
\begin{subfigure}{0.43\textwidth}\centering\includegraphics[width=0.95\columnwidth]{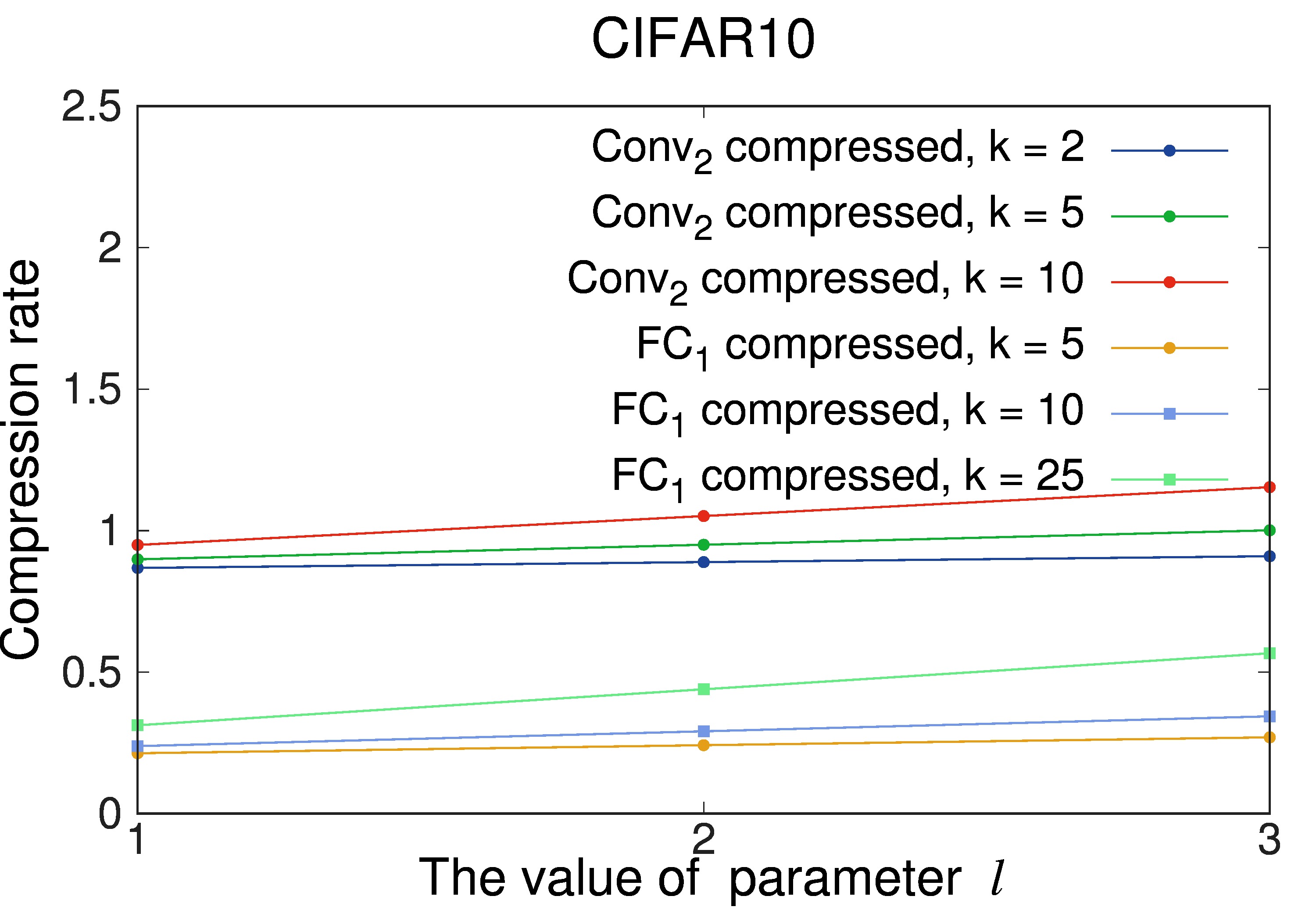}\end{subfigure}&
\begin{subfigure}{0.43\textwidth}\centering\includegraphics[width=0.95\columnwidth]{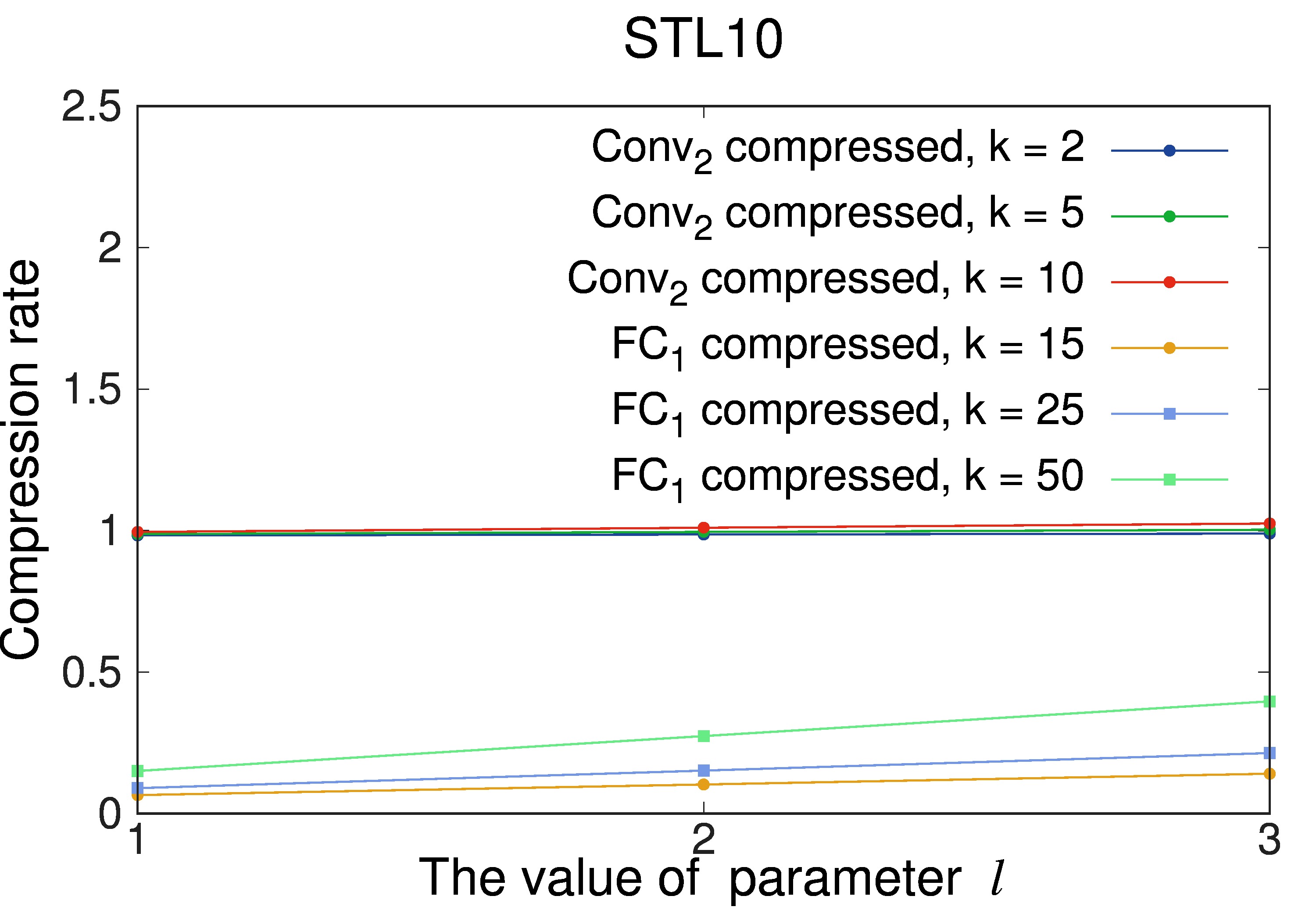}\end{subfigure} 
\end{tabular}
\caption{The plots on the left show the top-$1$ error and the compression rate for the CIFAR10 dataset obtained by varying $k$ and $\ell$ in our tensor sketching approach on TestNet. The plots on the right show the same for the STL10 dataset.
}
\label{t:diffparams}
\end{center}
\end{figure*}

\paragraph{Parameter Sensitivity.} For understanding the role of parameters $k$ and $\ell$ in our tensor sketching approach, we train a number of networks derived from the TestNet architecture for several combinations of these parameters. For the convolutional layer, we construct different networks each of which is obtained by replacing the $\Conv_2$ layer of TestNet with a \SK-\Conv layer for different values of $k$ and $\ell$. We vary $\ell \in \{1,2,3\}$ and $k \in \{2,5,10\}$, independently, giving rise to $9$ different networks. Similarly, we also construct new networks by replacing the $\FC_1$ layer of TestNet with a \SK-\FC layer with $\ell \in \{1,2,3\}$ and $k \in \{5,10,25\}$ for smaller images (like CIFAR10) and $k \in \{15,25,50\}$ for larger images (like STL10). Figure~\ref{t:diffparams} shows the results for the CIFAR10 and STL10 datasets (results on other two datasets are similar and omitted here). For each compressed model, we show its top-$1$ error (plots in the top row). The plots in the bottom row present the corresponding compression rate for each network. Note that if the parameters $k$ and $\ell$ are set too high then the compression rate can be $> 1$. In this case, we have an expansion over the original network. If a point is missing from a line then the corresponding network failed in the training. As an example, consider the network obtained by replacing $\FC_1$ layer with a \SK-\FC layer using $k=5$ and $\ell=2$. From the plot in Figure~\ref{t:diffparams}, the model obtained by training this network on CIFAR10 has $\topone \approx 30\%$. We also see that this network has a compression rate of $\approx 0.5$, i.e., the size of the original TestNet has been reduced by a factor of 2. Recall that by design TestNet has much more parameters in the fully connected layers than the convolutional layers, hence compressing the $\FC_1$ layer leads to smaller compression rates than compressing the $\Conv_2$ layer (as observed in Figure~\ref{t:diffparams}).


First, from Figure~\ref{t:diffparams}, we observe that the accuracy of the compressed models improve as we increase $k$ or $\ell$. This is expected because, as we discuss in Section~\ref{sec:architecture}, by increasing $k$ or $\ell$ we are increasing the effective size of the constructed sketches. For example, on the STL10 dataset with $\ell=1$ and the fully connected layer compression, as we increase $k$ from $15$ to $50$, the top-$1$ error goes down from around $62\%$ to $45\%$ (which is comparable to the $40.5\%$ top-$1$ error of the original (uncompressed) model from Table~\ref{t:base}). However, with increasing $k$ or $\ell$ the compression rate goes up (implying lower overall compression).

Second, due to the averaging effect, increasing $\ell$ increases the stability in the sketching process. For example, consider CIFAR10 where $\FC_1$ layer is replaced with a \SK-\FC layer using $k=10$ and $\ell\in \{1,2,3\}$. We trained each resulting network architecture $10$ different times each time initializing the \SK-\FC layer with a different set of random matrices $U_{1_1},\dots,U_{1_\ell},U_{2_1},\dots,U_{2_\ell}$ and measured the variance in the top-$1$ error across different runs. Not surprisingly, increasing $\ell$ decreases the variance of top-$1$ error with respect to the randomization that arises from the use of random matrices. For example, with $\ell=1$ we get an average (over these runs) top-$1$ error of $30.1$ with variance of $0.44$, for $\ell=2$ we get an average top-$1$ error of $29.1$ with variance of $0.29$, and for $\ell=3$ we get an average top-$1$ error of $28.6$ with variance of $0.23$.


\end{document}